\documentclass[10pt,twocolumn,letterpaper]{article}

\usepackage[pagenumbers]{cvpr} 

\usepackage{times}
\usepackage{epsfig}
\usepackage{graphicx}
\usepackage{amsmath}
\usepackage{amssymb}
\usepackage{caption}

\usepackage{booktabs}
\usepackage{color}
\usepackage[accsupp]{axessibility}
\usepackage{tabularx}
\usepackage{boxhandler}
\usepackage{adjustbox}
\usepackage{multirow}
\usepackage{verbatim}
\usepackage{amssymb}
\usepackage{pifont}
\usepackage{lipsum}

\usepackage[pagebackref,breaklinks,colorlinks]{hyperref}

\usepackage[capitalize]{cleveref}
\crefname{section}{Sec.}{Secs.}
\Crefname{section}{Section}{Sections}
\Crefname{table}{Table}{Tables}
\crefname{table}{Tab.}{Tabs.}

\def\cvprPaperID{6292} 
\def\confName{CVPR}
\def\confYear{2022}

\begin{document}

\title{Styleformer: Transformer based Generative Adversarial Networks with Style Vector}

\author{Jeeseung Park$^*$\\
mAy-I Inc.\\
{\tt\small jspark@may-i.io}
\and
Younggeun Kim\thanks{Equal contribution. The order was determined randomly.}\\
MINDsLab Inc.\\
{\tt\small younggeun@mindslab.ai}
}
\maketitle

\begin{abstract}
  
 We propose Styleformer, a generator that synthesizes image using style vectors based on the Transformer structure. 
In this paper, we effectively apply the modified Transformer structure (e.g., Increased multi-head attention and Pre-layer normalization) and introduce novel Attention Style Injection module which is style modulation and demodulation method for self-attention operation.
The new generator components have strengths in CNN's shortcomings, handling long-range dependency and understanding global structure of objects. 
We present two methods to generate high-resolution images using Styleformer. 
First, we apply Linformer in the field of visual synthesis (Styleformer-L), enabling Styleformer to generate higher resolution images and result in improvements in terms of computation cost and performance. This is the first case using Linformer to image generation.
Second, we combine Styleformer and StyleGAN2 (Styleformer-C) to generate high-resolution compositional scene efficiently, which Styleformer captures long-range dependencies between components. 
With these adaptations, Styleformer achieves comparable performances to state-of-the-art in both single and multi-object datasets. Furthermore, groundbreaking results from style mixing and attention map visualization demonstrate the advantages and efficiency of our model.

\end{abstract}

\section{Introduction}
\label{sec:intro}

Generative Adversarial Network (GAN) \cite{goodfellow2014generative} is one of the widely used generative model. Since the appear of DCGAN \cite{radford2016unsupervised}, convolution operations have been considered essential for high-resolution image generation and stable training. Convolution operations are created under the assumption of the locality and stationarity of the image (i.e., inductive bias), which is advantageous for image processing \cite{NIPS2012_c399862d}. 
Through convolution neural networks (CNNs) with this strong inductive bias, GAN have efficiently generated realistic, high-fidelity images.

However, drawbacks of CNNs clearly exist. 
 Local receptive field of CNNs makes model difficult to capture long-range dependency and understanding global structure of object. Stacking multiple layers can solve this problem, but this leads to another problem of losing spatial information and fine details \cite{yu2018generative}. Moreover, sharing kernel weights across locations leads to unstable training when the pattern or styles differ by location in the image \cite{zhang2019selfattention}. 
 This is also related to the poor quality of generated structured images or compositional scenes (e.g., outdoor scenes), unlike the generation of a single object (e.g., faces)


In this paper, we propose Styleformer, a generator that uses style vectors based on the Transformer structure. Unlike CNNs, Styleformer utilizes self-attention operation to capture long-range dependency and understand global structure of objects efficiently. Furthermore, we overcome computation problem of Transformer and show superior performance not only in low-resolution but also in high resolution images. Specifically, we introduce the following three models:

1) \textit{Styleformer} - The basic block of Styleformer is based on Transformer encoder, so we introduce components that need to be changed for stable learning. Inspired by MobileStyleGAN \cite{belousov2021mobilestylegan}, we enhance the multi-head attention in original Transformer by increasing the number of heads, allowing model to generate image efficiently. We also modify layer normalization, residual connection, and feed-forward network (Section \ref{section:3-2}). Moreover, we introduce novel \textit{attention style injection} module, suitable style modulation, and demodulation method for self-attention operation (Section \ref{section:3-3}). This design allows Styleformer to generate image stably, and enables model to handle long-range dependency and understand global structures.

2) \textit{Styleformer-L} - We sidestep scalability limitation arising from the quadratic mode of attention operation by applying Linformer \cite{wang2020linformer} 
(Styleformer-L). As such, Styleformer-L can generate high-resolution images with linear computational costs. 
This paper is the first case to apply Linformer in the field of visual synthesis (Section \ref{section:3-4}). 

3) \textit{Styleformer-C} - We further combine Styleformer and StyleGAN2, applying Styleformer at low resolution and style block of StyleGAN2 at high resolution (Styleformer-C). As can be seen from our experiments and analysis (e.g., style mixing and visualizing attention map), we show that Styleformer-C with the structure above can generate compositional scenes efficiently, and showing flexibility of our model. In detail, we prove that Styleformer in low resolution help model to capture long-range dependency between components, and style block in high resolution help model to refine the details of each components such as color or texture. This novel blending structure enables fast training, which is the advantage of StyleGAN2, while maintaining the advantages of Styleformer that can generate structured images.(Section \ref{sec:4}). 

Styleformer achieves comparable performances to state-of-the-art in both single and multi-object datasets.
We record FID 2.82 and IS 10.00 at the unconditional setting on CIFAR-10. These results outperform all GAN-based models including StyleGAN2-ADA \cite{karras2020training} which recently recorded state-of-the-art. As can be expected, Styleformer show strength especially in multi-object images or compositional scenes generation (e.g., CLEVR, Cityscapes).
Styleformer-C records FID 11.67, IS 2.27 in CLEVR, and FID 5.99, IS 2.56 in Cityscapes, showing better performance than pure StyleGAN2. 
\section{Related Work}
After origion of GAN \cite{goodfellow2014generative}, various methods  \cite{arjovsky2017wasserstein, mao2017squares,miyato2018spectral,karras2018progressive} have been proposed to enhance its training stability and performance. As a result, fidelity and diversity of the generated images have dramatically improved. 
In addition to image synthesis task, GAN has been widely adopted in various tasks, such as image-to-image translation \cite{isola2018imagetoimage, zhu2020unpaired}, super resolution \cite{ledig2017photorealistic}, image editing \cite{yu2018generative}, and style transfer \cite{choi2018stargan}. In particular, StyleGAN-based architectures have been applied for various applications \cite{gabbay2019style,zhu2020semantically,Zhu_2020}. However, since all of these models are based on convolution backbones, they have met with only limited success on generating complex or compositinal scenes \cite{johnson2018image}. 

Transformer \cite{vaswani2017attention} was first introduced to the natural language processing(NLP) domain, achieving a significant advance in NLP. Recently, there were efforts to utilize Transformer in the computer vision field \cite{dosovitskiy2020image, bertasius2021spacetime, zheng2021rethinking}. 
Using huge amounts of data and a transformer module, ViT \cite{dosovitskiy2020image} obtains comparable result with state-of-the-art model in the existing CNN based image classification model \cite{tan2020efficientnet,kolesnikov2020big}. Inspired by \cite{dosovitskiy2020image}, various models such as \cite{graham2021levit,wu2021cvt,liu2021swin} emerges based on this structure. There have also been attempts to utilize transformer for tasks such as video understanding \cite{bertasius2021spacetime} and segmentation \cite{zheng2021rethinking} as well as image classification. 
Even in GAN, there have been attempts to utilize Transformer: GANformer \cite{hudson2021generative} proposes a bipartite Transformer structure and applies it to StyleGAN \cite{karras2019stylebased,karras2020analyzing}. With this structure, GANformer successfully advance the generative
modeling of structured images and scenes, which have been challenging in existing GAN. However, they use a bipartite attention, differ from the self-attention operation.
TransGAN\cite{jiang2021transgan} demonstrates a convolution-free generator based on the structure of vanilla GAN, which doesn’t show good performance compared to state-of-the-art model. 

\begin{figure}[t]
\begin{center}

\includegraphics[width=1.\linewidth]{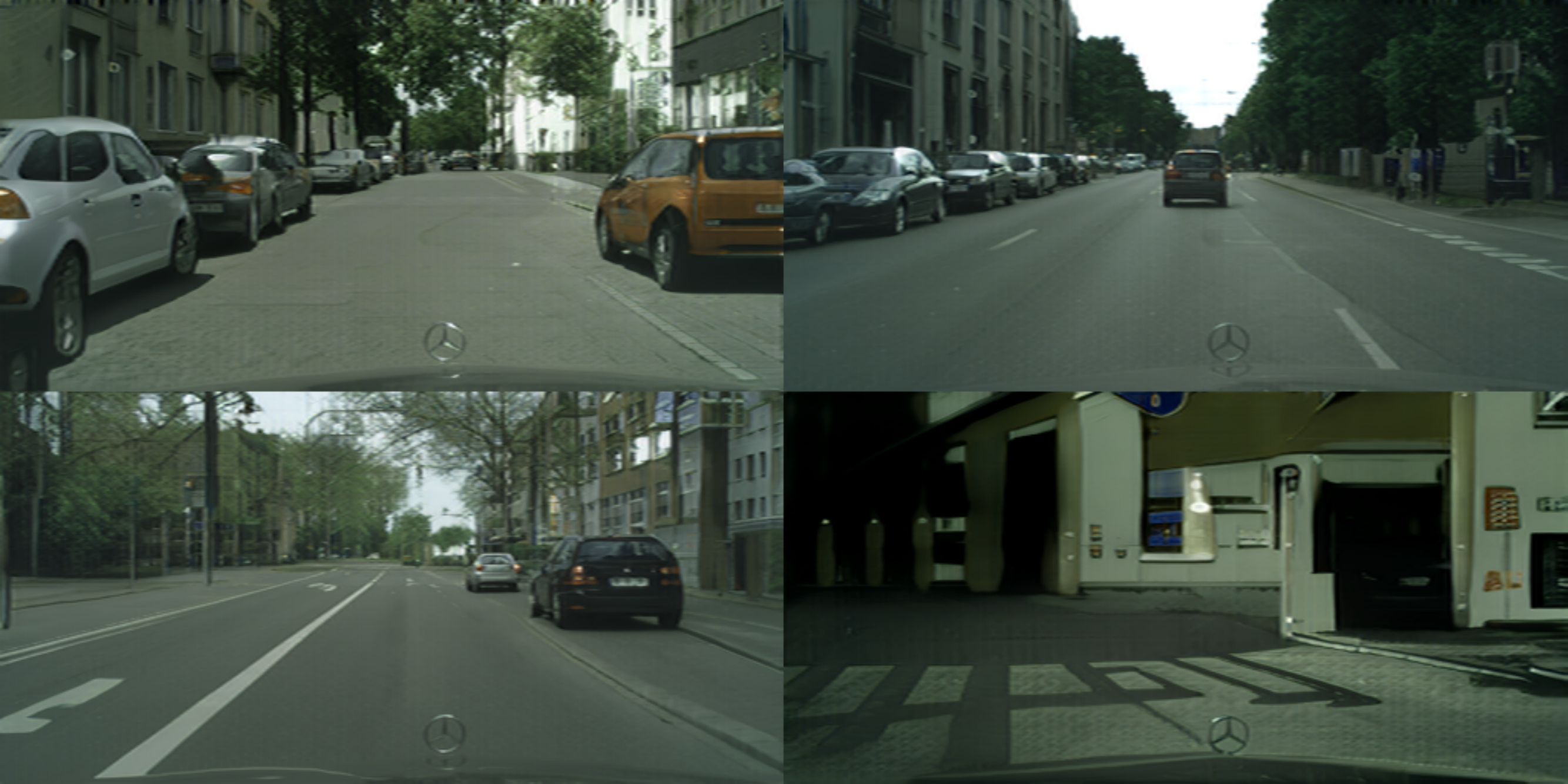}

\end{center}
\vspace{-5mm}
   \caption{High-resolution compositional scenes generated by Styleformer.
}
\label{fig:afhq}
\end{figure}

Unlike these studies, Styleformer generate images with self-attention operation using style vector and showing comparable performance state-of-the-art models \cite{karras2019stylebased,karras2020analyzing}. Previous methods (TransGAN) mainly use pre-defined sparse attention patterns for efficient attention mechanism, but we explore the low-rank property in self-attention. Our model can generate high resolution images ($512 \times 512$) with reduced computation complexity, while GANformer and TransGAN show a maximum of $256 \times 256$ image synthesis.

\begin{figure*}[t]
\begin{center}

\includegraphics[width=.7\linewidth]{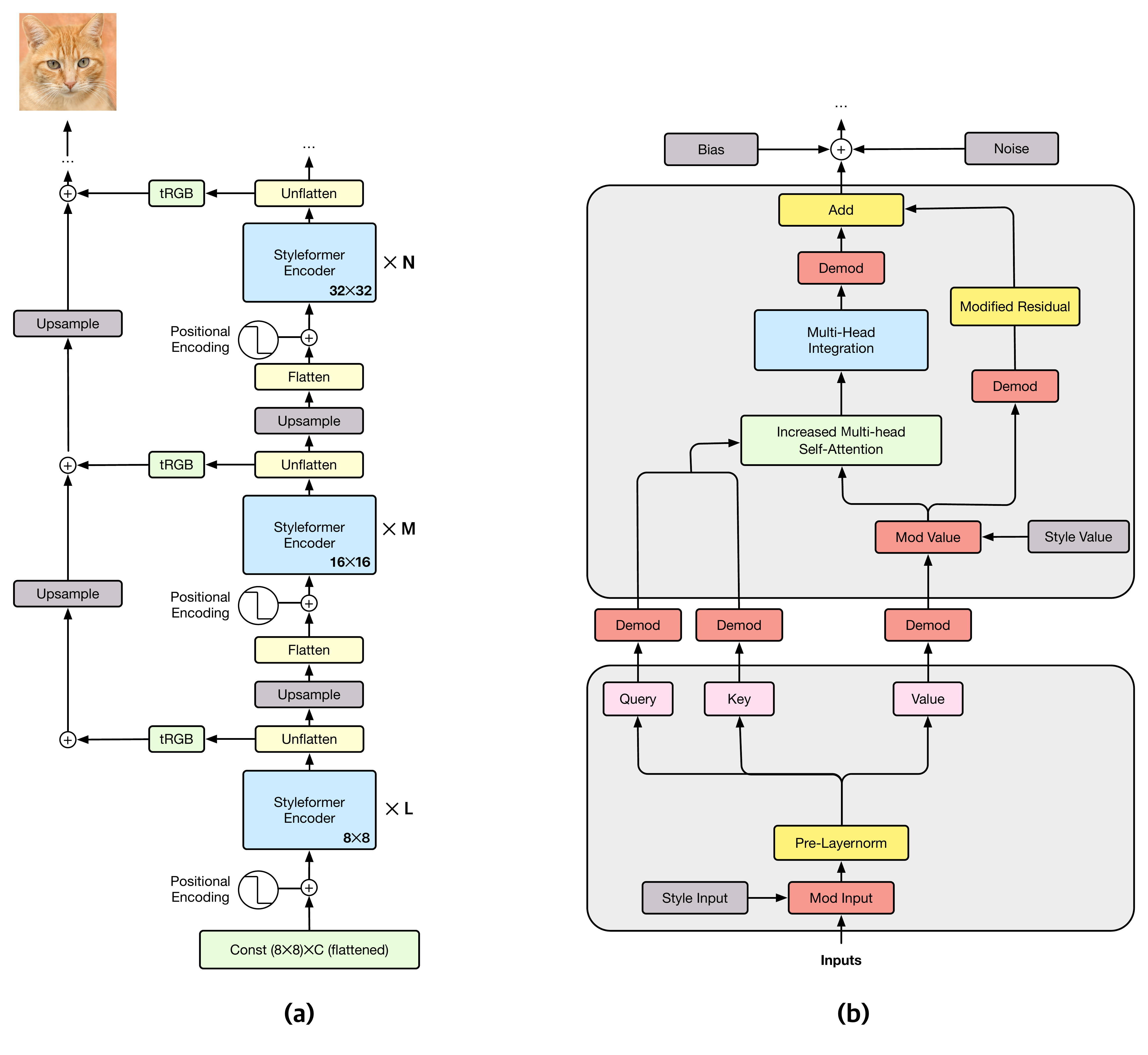}

\end{center}
\vspace{-5mm}
   \caption{(a) Overall Architecture of Styleformer. (b) Styleformer encoder structure, which is the basic block of Styleformer.
}
\vspace{-3mm}
\label{fig:overall}
\end{figure*}

\section{Styleformer}
\label{sec:3}


\subsection{Styleformer Architecture}
\label{section:3-1}

Figure \ref{fig:overall}a shows the overall architecture of Styleformer, and in Figure \ref{fig:overall}b we show Styleformer encoder network, the basic block of Styleformer. Like existing synthesis network of StyleGAN, our generator is conditioned on a learnable constant input. The difference is that the constant input ($8 \times 8$) is flattened (64) to enter the Transformer-based encoder. Then the input which is combined with learnable positional encoding passes through the Styleformer encoder. Styleformer encoder is based on Transformer encoder, but there are several changes to generate an image efficiently, which will be discussed in Section \ref{section:3-2}.  

After passing several encoder blocks in each resolution, we proceed bilinear upsample operation by reshaping encoder output to the form of a square feature map. After upsampling, flatten process is carried out again to match the input form of the Styleformer encoder. This process is repeated until the feature map resolution reaches the target image resolution. For each resolution, the number of the Styleformer encoder and hidden dimension size can be chosen as hyperparameters.


\subsection{Styleformer Components from Transformer}
\label{section:3-2}



\paragraph{Increased Multi-Head Attention}

Modern vision architectures allow communications between different channels and different spatial locations (i.e., pixels) \cite{tolstikhin2021mlpmixer}. Conventional CNNs perform the above two communications at once, but these communications can be clearly separated like depthwise separable convolutions \cite{howard2017mobilenets}. We also separate the pixel-communication (self-attention), channel-communication operations (multi-head integration) in the Transformer encoder. However, in depthwise separable convolutions, distinct convolution kernels are applied to each channel, unlike the self-attention operation share only one huge kernel $A$ (i.e., attention map). With same kernel applied to each channel, diversity in generated image can be decreased.

\begin{figure}[t]
\begin{center}

\includegraphics[width=.45\linewidth]{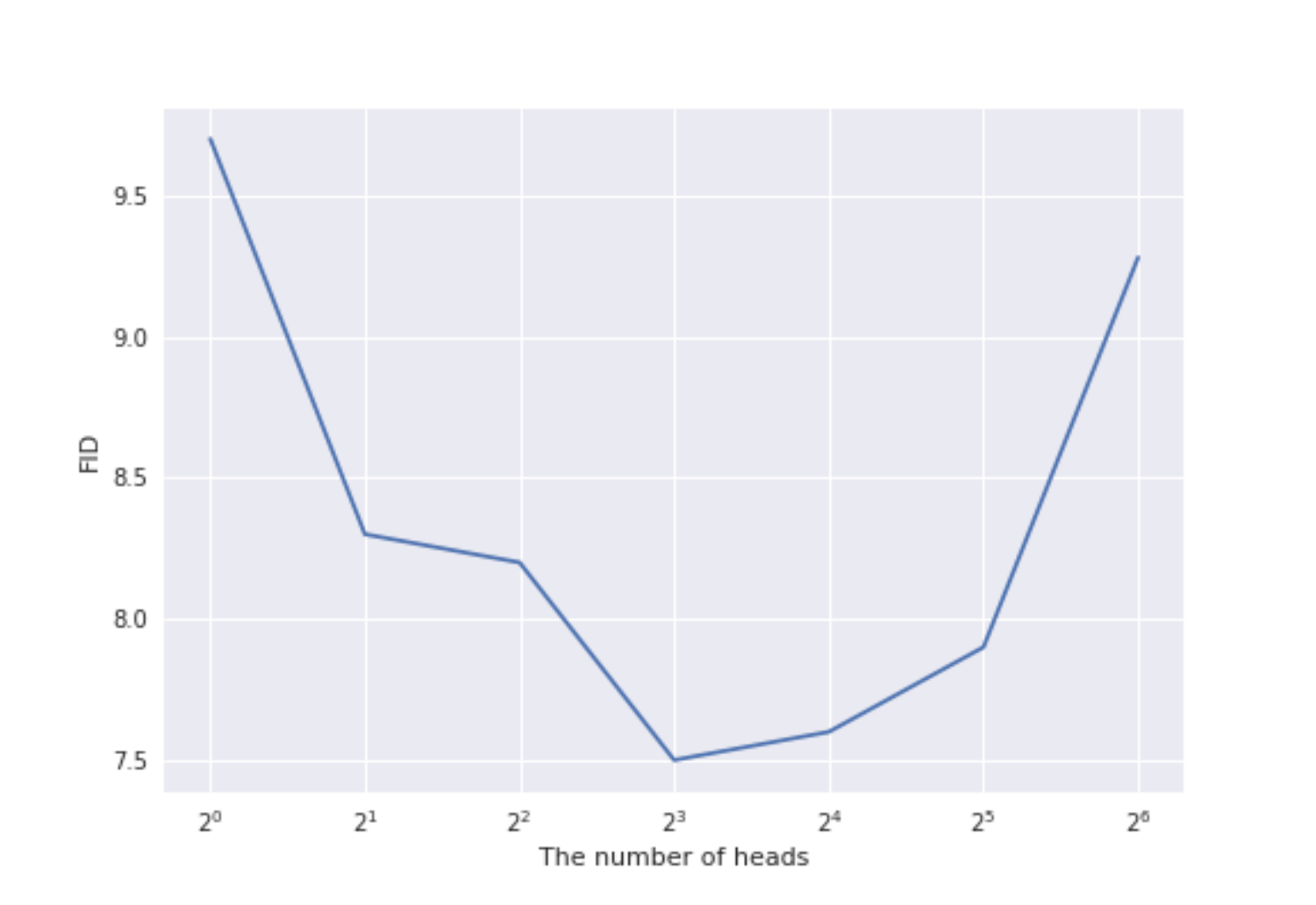}
\includegraphics[width=.45\linewidth]{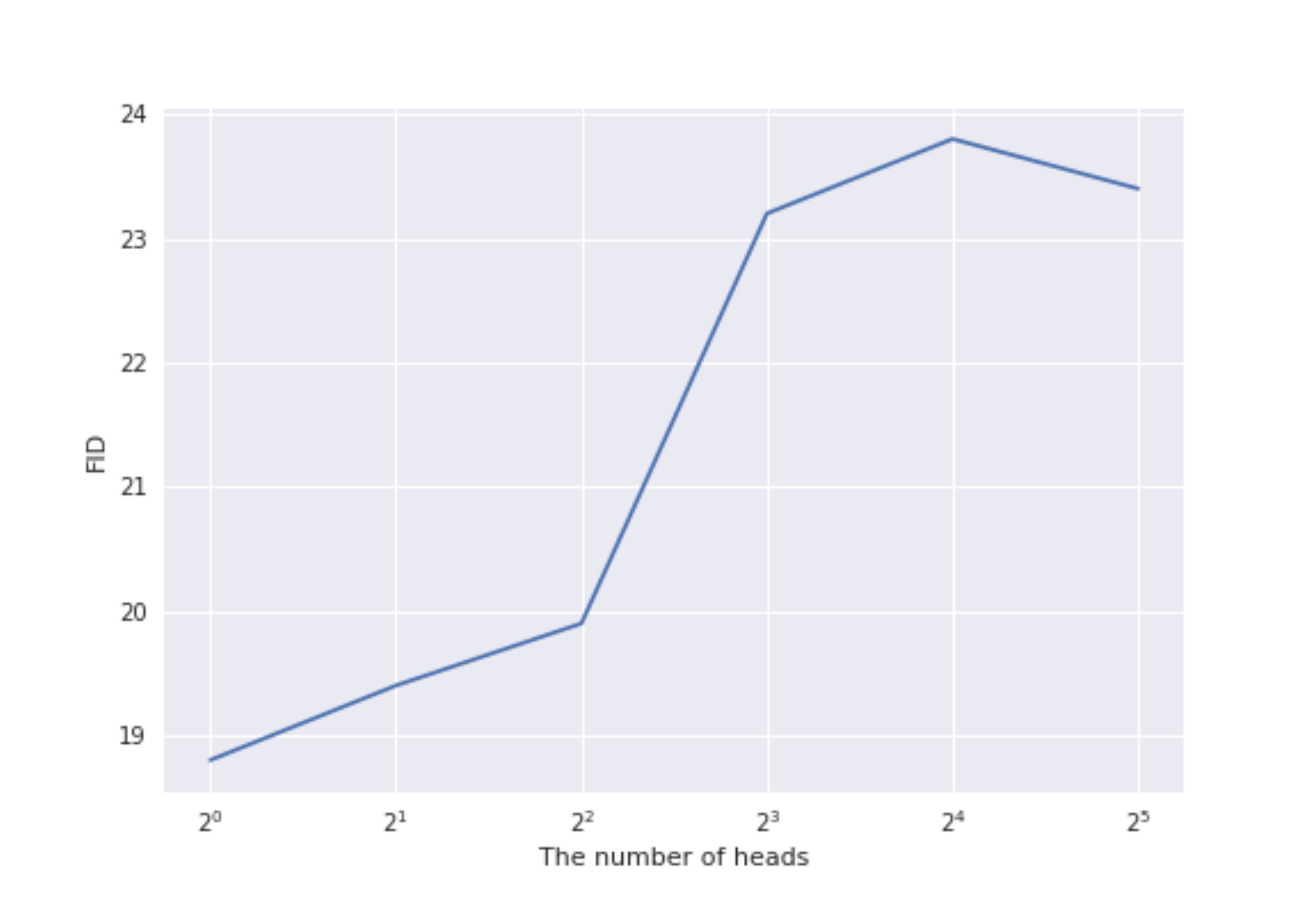}
\end{center}
\vspace{-5mm}
   \caption{It shows FID on CIFAR-10 with one layer Styleformer, which hidden dimension size is fixed as 256 and 32, respectively. Both experiments show the best result when the depth is 32.
}
\vspace{-3mm}
\label{fig:multihead_graph}
\end{figure}

We overcome this problem by increasing the number of heads of multi-head attention (Increased multi-head attention). Then, the created attention map will be different for each head, and so the kernel applying operation. 
Then the attention maps will be created for each head, making the channels in each head meet different kernels.
However, increasing the number of heads too much may cause attention map to not be properly created, resulting in poor performance.
We demonstrate experimentally that increasing the number of heads improves performance only when the depth is at least 32, as shown in Figure \ref{fig:multihead_graph}. 
Therefore, we fix the depth to 32 for all future experiment.  
More details about increased multi-head attention can be found in Appendix $C$. 

\begin{figure}[t]
\begin{center}

\includegraphics[width=.8\linewidth]{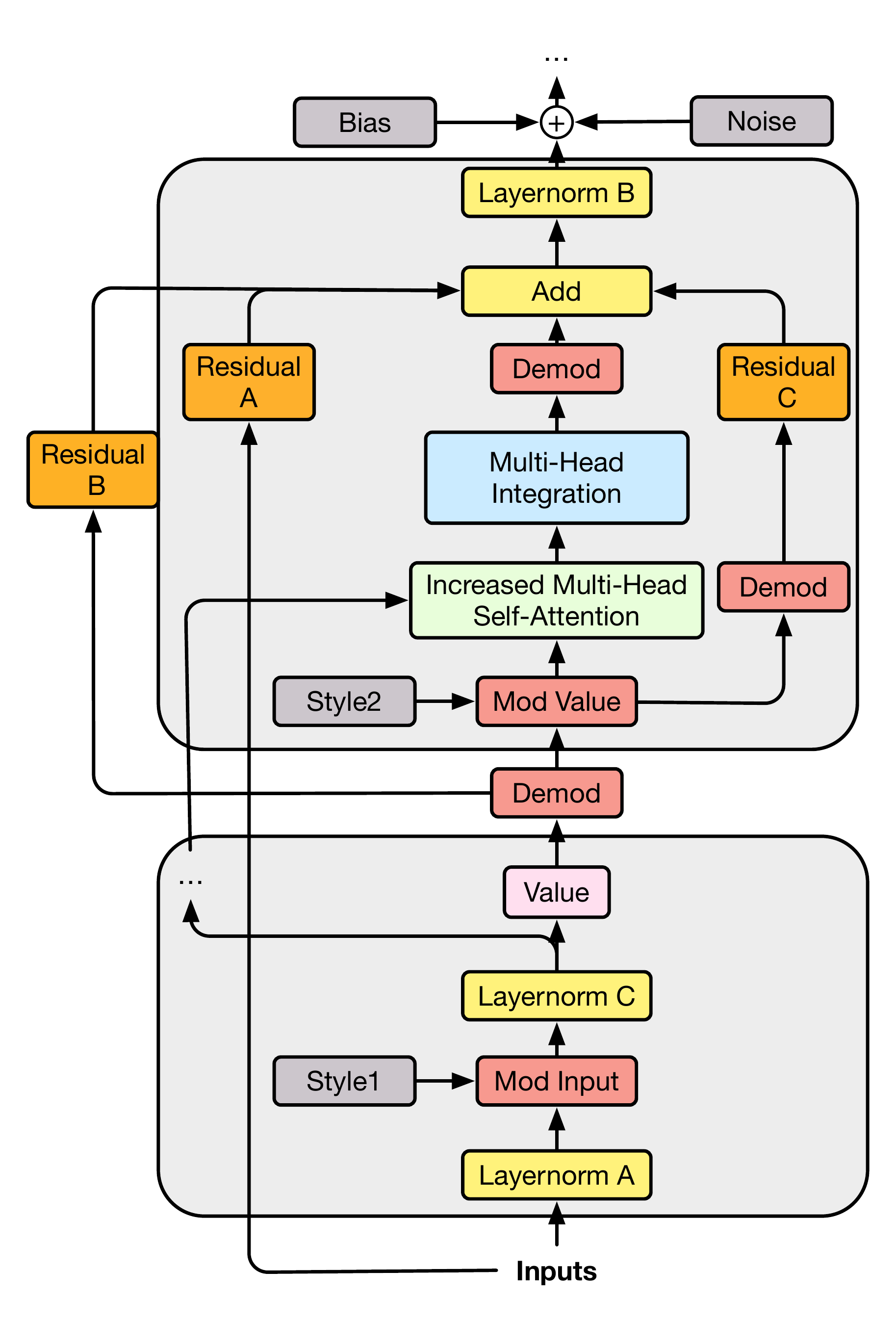}
\end{center}
\vspace{-5mm}
   \caption{Styleformer encoder structure for ablation study, including residual connection, layer normalization, attention style injection.
}
\vspace{-3mm}
\label{fig:ablation}
\end{figure}

\paragraph{Pre-Layer Normalization}
We change the position of layer normalization in Transformer encoder. The layer normalization of the existing Transformer comes after a linear layer that integrates multi-heads (Post-Layer normalization). We hypothesis that the role of layer normalization in a Transformer is the preparation of generating an attention map. If we perform layer normalization at the end of Styleformer encoder (\textit{Layernorm B} in Figure \ref{fig:ablation}), style modulation is applied before making query and key, which can disturb learning attention map. This is supported by ablation study and attention map analysis in Table \ref{table_ablation:1} and Appendix $B$, respectively. 
Therefore, to solve this problem, we proceed layer normalization before operation making query, key and value (\textit{Pre-Layernorm} in Figure \ref{fig:overall}b) 

\paragraph{Modified Residual Connection}
Unlike the Transformer encoder, input feature map is scaled by style vector (\textit{Mod input} in Figure \ref{fig:overall}b) in Styleformer encoder. We hence find the residual connection suitable for scaled input. After ablation study, we apply residual connection like \textit{Modified Residual} in Figure \ref{fig:overall}b. Demodulation operation is additionally performed in residual connection, which will be described in Section \ref{section:3-3}. Table \ref{table_ablation:1} presents ablation details of residual connection. 

\paragraph{Eliminating Feed-Forward Network}

As can be seen Table \ref{table_ablation:1}, we remove the feed-forward structure because eliminating feed forward structure makes the model perform better and more efficient.

\begin{table*}[ht]
\centering
\resizebox{1.\linewidth}{!}{
\begin{tabular}{c|ccccccccccc}
\noalign{\smallskip}\noalign{\smallskip}\hline\hline
\multirow{2}{*}{\textbf{Method}} & \multirow{2}{*}{Style1} & \multirow{2}{*}{Style2} & \multirow{2}{*}{Style1=Style2} & \multirow{2}{*}{Residual A} & \multirow{2}{*}{Residual B} & \multirow{2}{*}{Residual C} & \multirow{2}{*}{Layernorm A} & \multirow{2}{*}{Layernorm B} & \multirow{2}{*}{Layernorm C} & \multirow{2}{*}{Feed-Forward} &\multirow{2}{*}{FID}  \\
& & & & & & & & & & & \\
\hline
\midrule
\multirow{1}{*}{\textbf{Baseline}}& O & O & X & X & X & O & X & X & O & X& 8.56\\
\midrule
\multirow{3}{*}{\textbf{Attention Style Injection}}&  O & X & - & X & X & O & X & X & O & X& 11.01\\
&  X & O & - & X & X & O & X & X & O & X &  11.40\\
&  O & O & O & X & X & O & X & X & O & X& 10.27\\
\midrule
\multirow{3}{*}{\textbf{Residual Connection}} & O & O & X & X & X & X & X & X & O & X& 19.09\\
&  O & O & X & O & X & X & X & X & O & X&  14.70\\
& O & O & X & X & O & X & X & X & O & X& 9.94\\
\midrule
\multirow{2}{*}{\textbf{Layer Normalization}}& O & O & X & X & X & O & O & X & X & X& 9.00\\
& O & O & X & X & X & O & X & O & X & X&  10.96\\
\midrule
\multirow{1}{*}{\textbf{Feed-Forward}}& O & O & X & X & X & O & X & X & O & O& 14.75\\
\hline
\end{tabular}
}
\caption{Ablation details of Styleformer components. Ablation study was conducted using small version of Styleformer with CIFAR-10 dataset, trained for 20M images. See Appendix $A$ for further implementation details.}
\label{table_ablation:1}
\end{table*}

\subsection{Attention Style Injection}
\label{section:3-3}
Unlike vanilla GAN, StyleGAN generates an image with layer-wise style vectors as inputs, enabling controllable generation via style vectors, i.e., scale-specific control. Specifically, style vector scales the input feature map for each layer, i.e., style modulation, amplifying certain feature maps. For scale-specific control, this amplified effect must be removed before entering the next layer. StyleGAN allows scale-specific control through a normalization operation called AdaIN operation \cite{huang2017arbitrary, dumoulin2017learned, ghiasi2017exploring, dumoulin2018feature-wise}, which normalizes each feature map separately, then scale and bias each feature map with style vector. StyleGAN2 is an advanced form of StyleGAN and addresses the artifact problem caused by the AdaIN operation, solving it by demodulation operation. 
While the AdaIN operation normalize the output feature map directly, demodulation operation is based on statistical assumptions about the input feature map. For details, similar to the goal of normalization operation, demodulation operation aims to have an output feature map with a unit standard deviation while assuming that the input feature maps have a unit standard deviation, i.e., statistical assumption. Our goal is to design a Transformer-based generator that generates images through style vector while enabling scale specific control. Therefore, we propose style modulation, demodulation method for the self-attention operation, i.e., \textit{Attention style injection}.
\paragraph{Modulation for Self-Attention}
Just as the input feature map is scaled by style vector in the style block of StyleGAN2, the input feature map in the Styleformer encoder is also scaled by style vector (\textit{Mod Input} in Figure \ref{fig:overall}b).
 But unlike convolution operation in StyleGAN2, there are two steps in self-attention operation: dot product of query and key to create an attention map (i.e. kernel), weighted sum of the value with calculated attention map.
We hypothesis that the style vector applied to the operation in each step should be different. 
Therefore, we perform style modulation twice as in Figure \ref{fig:overall}b (\textit{Mod Input}, \textit{Mod Value}). This hypothesis is supported in Table \ref{table_ablation:1}. 
In Figure \ref{fig:overall}b,  \textit{Style Input} is a style vector for input, and \textit{Style Value} is a style vector only for value. Two style vectors are created through common mapping networks as in StyleGAN but different learned affine transformations. 

\paragraph{Demodulation for Query, Key, Value} 
As shown in Figure \ref{fig:overall}b, Styleformer encoder creates query ($Q$), key ($K$), and value ($V$) through linear operation to the input feature map scaled with \textit{Style Input} vector. After that, $V$ will be modulated with \textit{Style Value} vector additionally, so the demodulation operation for removing scaled effect of \textit{Style Input} is clearly required. Also, we observe that when an attention map is created with $Q$, $K$ from input scaled by \textit{Style Input}, specific value in the attention map becomes very large, demonstrated in Appendix \ref{app:B}. This prevents the attention operation from working properly. We sidestep this problem with demodulation operation to $Q$, $K$, before creating attention map. Eventually, demodulation operation is all required for $Q$, $K$, and $V$.

Let's first look at the style modulation to the input, i.e., \textit{Mod Input}. Each flattened input feature map is scaled through a style vector, which is equivalent to scaling the linear weight:
\begin{align}
    \label{eq:5}
    w'_{ij} &= s_i \cdot w_{ij},
\end{align}
where $w$ is original linear weight to make ($Q$, $K$, $V$) from flattened input feature map, and $w'$ is modulated linear weight. $s_{i}$ is $i$th component of style vector, which scales $i$th flattened input feature map, and $j$ means the dimension of ($Q$, $K$, $V$). 
Assuming that flattened input feature maps have unit standard deviation (i.e., statistical assumption of demodulation), after passing style modulation and linear operation, a standard deviation of output is as follows:
\begin{align}
    \label{eq:6}
    \sigma_{j} &= \sqrt{\sum_{i}{w^{'}_{ij}{^{2}}}}.
\end{align}
We scale output activations for each dimension of $Q$, $K$, and $V$ by $1/\sigma_{j}$(i.e., demodulation), making $Q$, $K$, and $V$ back to unit standard deviation. 

\paragraph{Demodulation for Encoder Output}
After demodulation operation to $Q$, $K$, and $V$, Styleformer encoder performs style modulation to $V$ (\textit{Mod Value}), weighted sum of $V$ with attention map (\textit{Increased Multi-head Self-attention}), and then performs linear operation (\textit{Multi-Head Integration}), as shown in Figure \ref{fig:overall}b. Encoder output will be input for next encoder, so demodulation operation is necessary.
We show in Appendix \ref{app:D} that, assuming $V$ has a unit standard deviation (This can be assumed because of demodulation for $V$), the standard deviation of Styleformer encoder output can be derived as follows:
\begin{align}
    \label{eq:7}
    \sigma^{'}_{lk} &= \sqrt{\sum_{\cdot}{A_{l\cdot}{^2}} \cdot \sum_{j}{w^{'}_{jk}{^2}}},
\end{align}
where $w'_{jk} = s_j \cdot w_{jk}$, i.e., modulated linear weight. $s_{j}$ scales $j$th feature map of $V$, and $k$ enumerates the flattened output feature map. Attention map $A$ is computed same as existing Transformer: dot products of $Q$ and $K$, divide each by square root of depth, and softmax function. $A_{l \cdot}$ denotes attention score vector for $l$th pixel. 

However, there are two problems with demodulation by simply scaling each flattened output feature map $k$ with $1/\sigma^{'}_{lk}$ (Equation \ref{eq:7}).
First, scaling output feature map $k$ with $1/\sigma^{'}_{lk}$ will normalize each pixel as a unit, different from AdaIN operation which normalizes each feature map as a unit.
Second, the attention map, which is a matrix derived from $Q$ and $K$, is dependent on the input. With input dependent variables, demodulation operations based on statistical assumptions can not be applied as in \cite{pmlr-v9-glorot10a}. Therefore we scale the flattened output feature map $k$ with $1/\sigma^{''}_{k}$ where
    $\sigma^{''}_{k} = \sqrt{\sum_{j}{w^{'}_{jk}{^2}}}$, normalizing each feature map as a unit, and excluding input dependent variables $A_{l}$. Then the standard deviation of output activations will be 
\begin{align}
    \sigma_{lk} = \cfrac{\sigma^{'}_{lk}}{\sigma^{''}_{k}} =  \sqrt{\sum_{\cdot}{A_{l\cdot}{^2}}}.
\end{align}
However in this way, standard deviation of output is not unit, rather approaching to zero when the numbers of pixels increase, as detailed in Appendix \ref{app:D}.
To prevent this effect, we have applied modified residual connection like \textit{Modified Residual} in Figure \ref{fig:overall}b. More specifically, we perform linear operation to \textit{Mod Value}, then perform demodulation operation (same as demodulation for query, key, value). With these modulation and demodulation operations in residual connection, variables with unit standard deviation are added to the output. Therefore it helps to keep the final output activation having unit standard deviation, when $\sigma_{jk}$ is close to zero.

\subsection{High Resolution Synthesis with Styleformer}
\label{section:3-4}
The main problem in applying Transformer to image generation is the efficiency problem with image resolution. In this section, we introduce two different techniques in Styleformer that can generate high resolution images. We show a method of applying Linformer, making computation complexity to linear. Then, we introduce a method of combining Styleformer and StyleGAN2, which can obtain the advantages of both models.

\paragraph{Applying Linformer to Styleformer}
For high-resolution images, input sequence length of the Styleformer encoder increases quadratically, and the standard self-attention mechanism requires a complexity of $O(n^2)$ with respect to the sequence length. It means attending to all pixels for each layer is almost impossible for high-resolution image generation. Therefore, we apply Linformer \cite{wang2020linformer} to our model, which projects key and value to the $k$ dimension when applying self-attention, reducing the time and space complexity from $O(n^2)$ to $O(nk)$. We fix $k$ to 256 and apply Linformer to the encoder block above $32 \times 32$ resolution, only when $n$ is 1024 or higher. We call this model as \textit{Styleformer-L}. 

\cite{wang2020linformer} explains that this new self-attention mechanism succeeds because the attention map matrix is low-rank. We observe this can be applied equally to the attention map matrix in the image: in the case of images, the pixel that needs to attend is often in a particular location, not all pixels(e.g. where objects are located in the image), which results in low-rank attention map matrix. Applying Linformer creates a more dense attention map, and also reduces computation. This is proved by spectrum analysis of attention map in Section \ref{section:4-2}. See Appendix \ref{app:E} for more details about Styleformer-L.  

\paragraph{Combining Styleformer and StyleGAN2}
Even with applying Linformer, it is difficult to generate an image for extremely high resolution like $512\times512$ using only Transformer. We solve this problem by combining Styleformer and StyleGAN2 to generate a high-resolution image, and we call this model \textit{Styleformer-C}. Styleformer-C is composed of Styleformer at low resolution, and style block of StyleGAN2 at high resolution. As demonstrated in \ref{section:4-1}, Styleformer encoder in low resolution help model to capture long-range dependency between components or global shape of object, and style block in high resolution help model to refine the details of each components or objects. In other words, model can capture global interactions efficiently using Styleformer only at low resolution, which leads to fast training speed. The overall architecture and details of Styleformer-C are described in Appendix \ref{app:F}.

\begin{figure*}[t]
\begin{center}
\includegraphics[height=.40\columnwidth, width=.40\columnwidth]{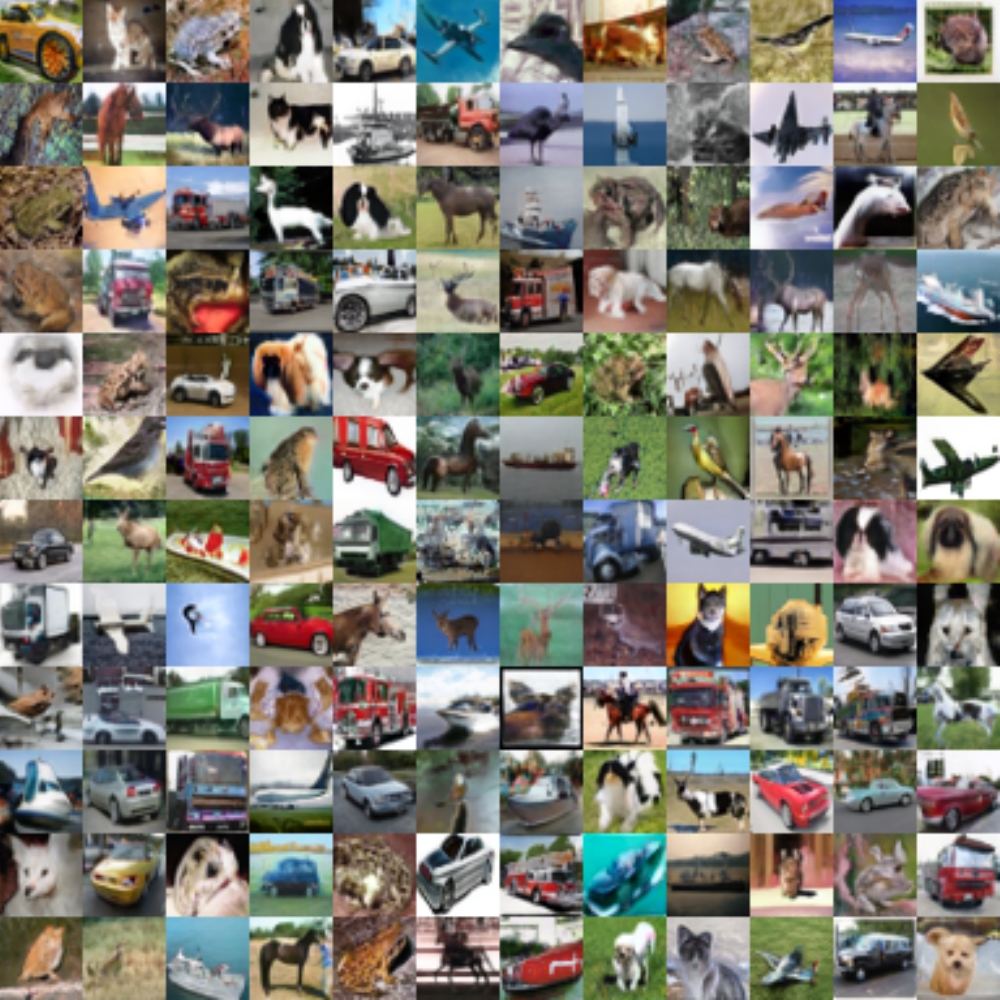}
\includegraphics[height=.40\columnwidth, width=.40\columnwidth]{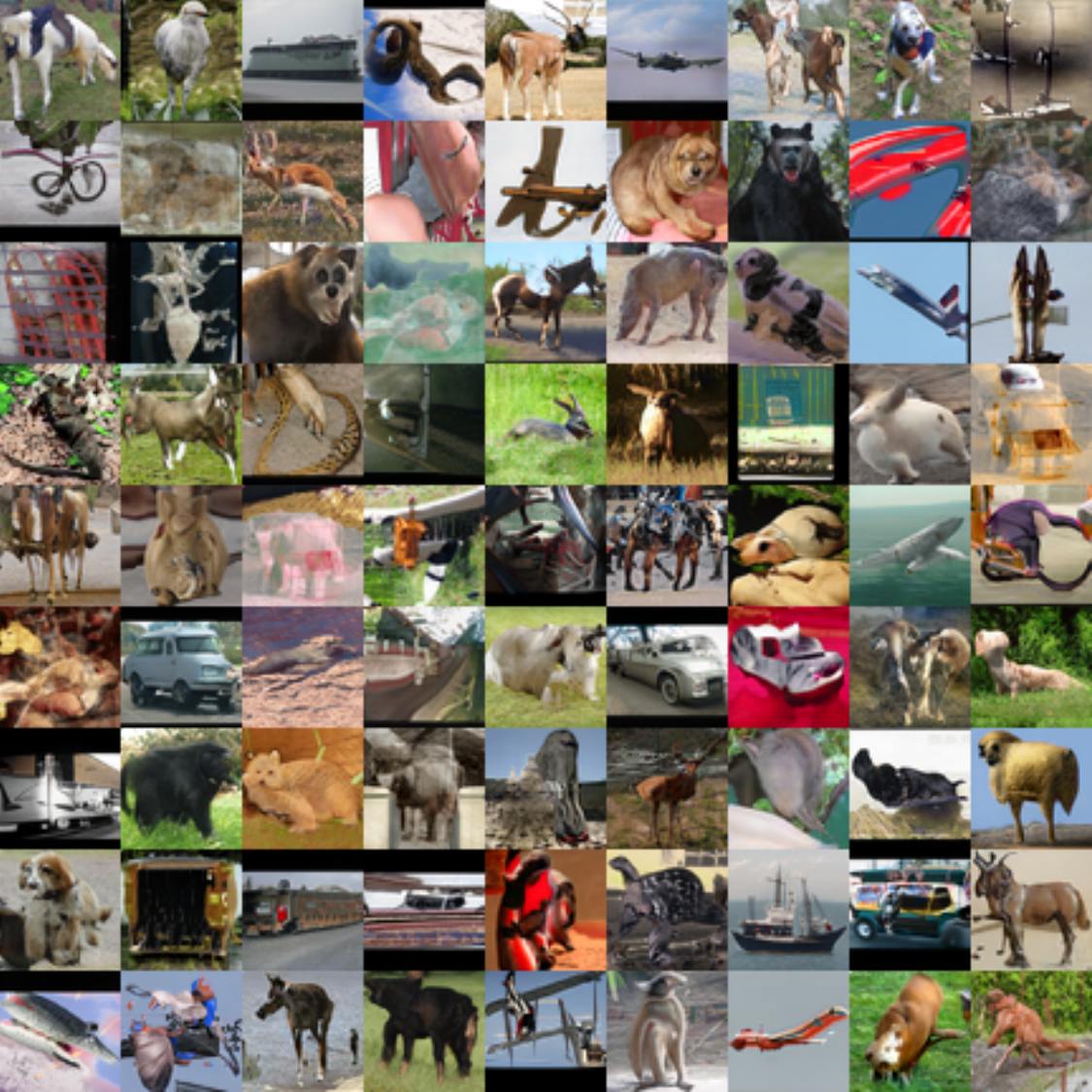}
\includegraphics[height=.40\columnwidth, width=.40\columnwidth]{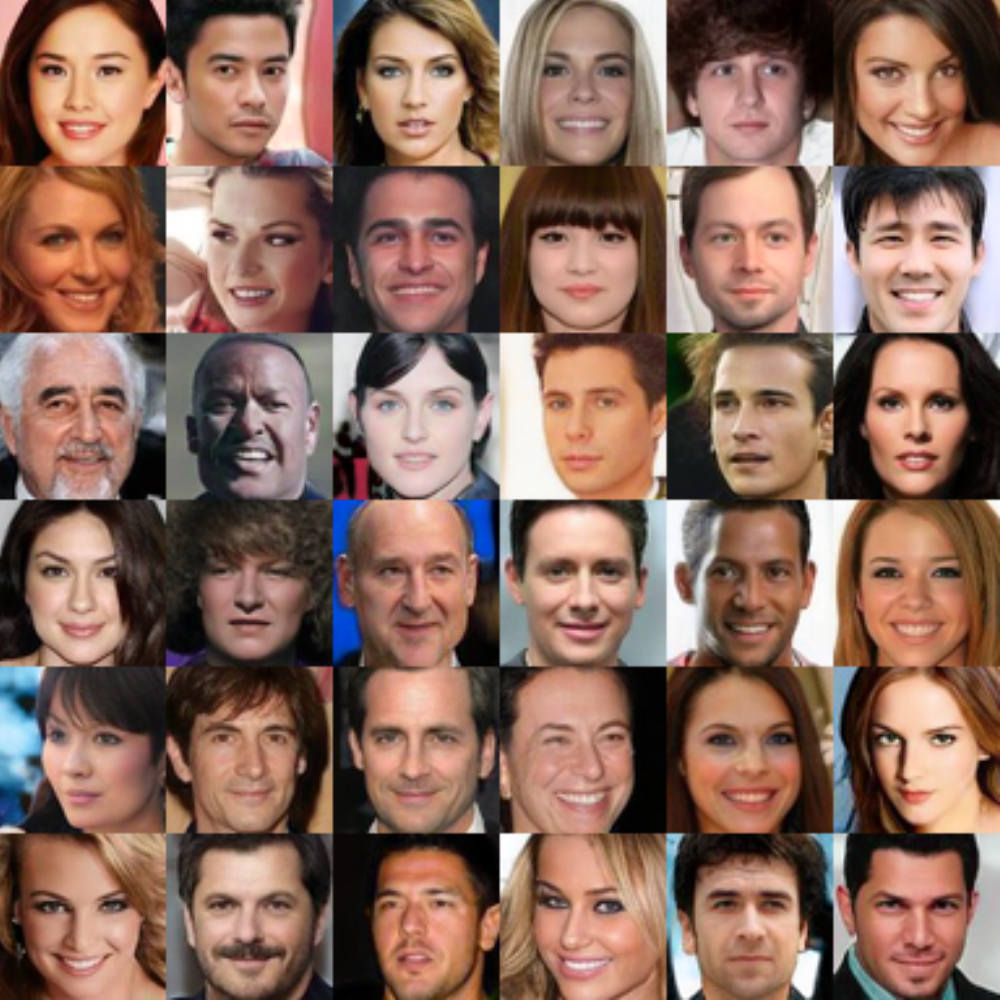}
\includegraphics[height=.40\columnwidth, width=.40\columnwidth]{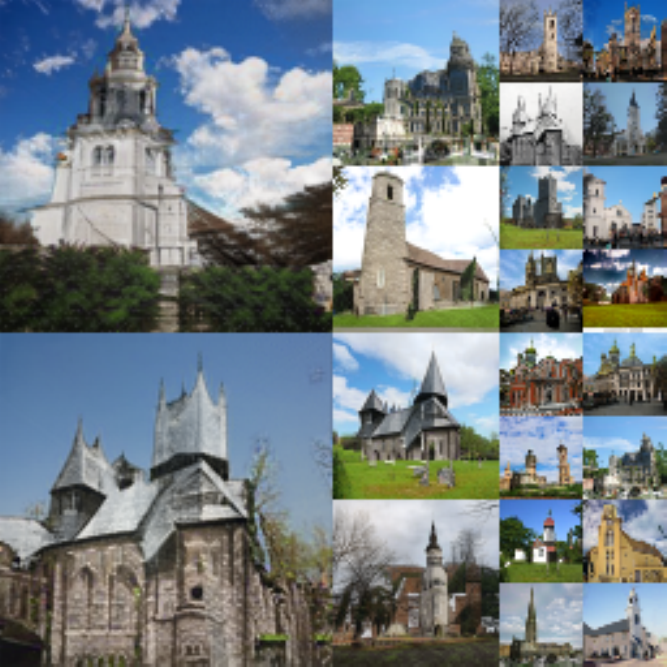}
\includegraphics[height=.40\columnwidth, width=.40\columnwidth]{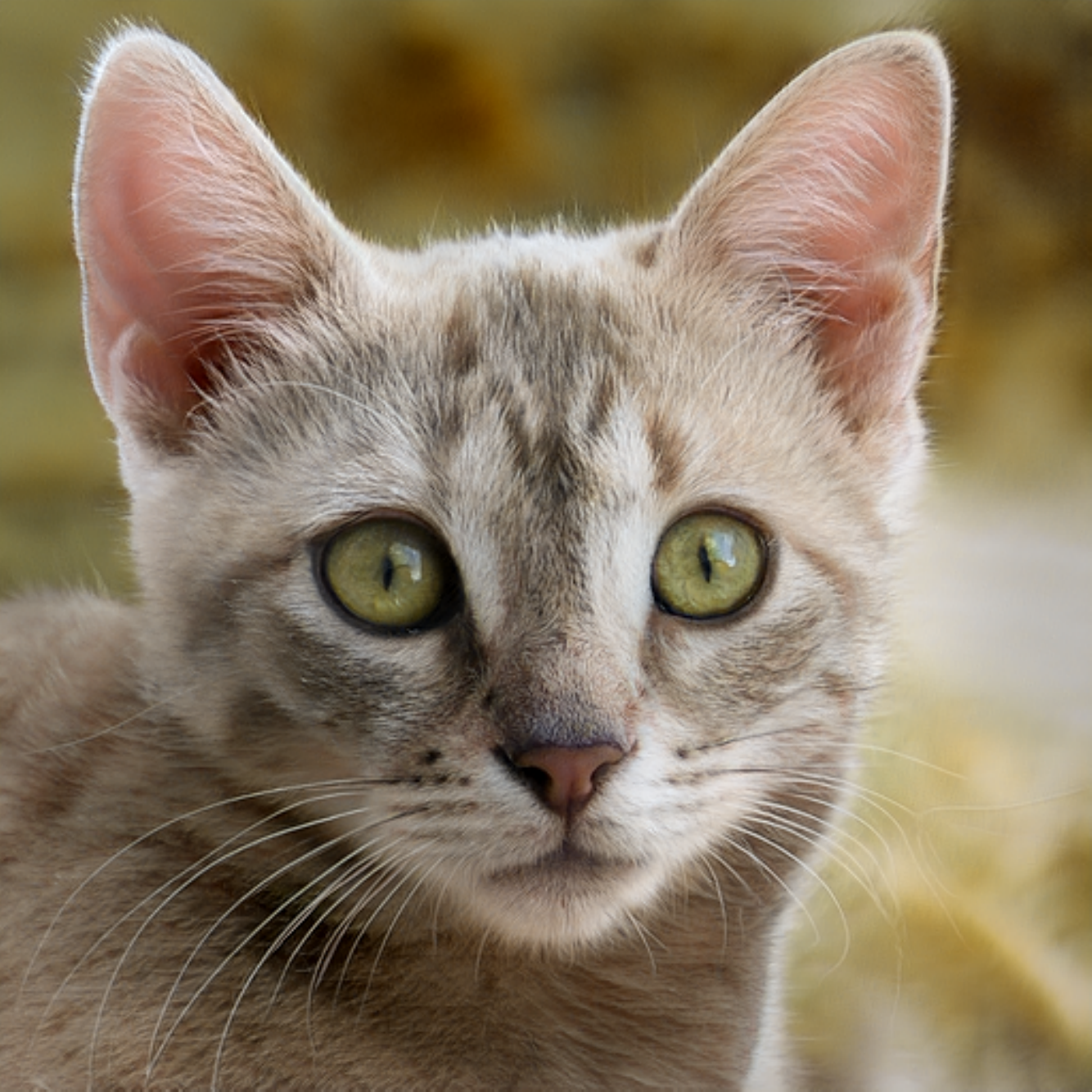}
\end{center}
\vspace{-5mm}
\caption{From the left, results generated by Styleformer on CIFAR-10 and STL-10, results generated by Styleformer-L on CelebA and LSUN-church and results generated by Styleformer-C on AFHQ-Cat. For more generated samples, please see Appendix \ref{app:G}.
}
\vspace{-3mm}
\label{fig:cifar}
\end{figure*}

\begin{table*}[!t]
\centering
\resizebox{1.\linewidth}{!}{
\begin{tabular}{ccc|ccc|cc}
\noalign{\smallskip}\noalign{\smallskip}\hline\hline
\multicolumn{3}{c|}{\textbf{Cifar-10}} & \multicolumn{3}{c|}{\textbf{STL-10}} &
\multicolumn{2}{c}{\textbf{CelebA}}\\
\multirow{1}{*}{\textbf{Method}} & \multirow{1}{*}{FID $\downarrow$}  & \multirow{1}{*}{IS $\uparrow$} &
\multirow{1}{*}{\textbf{Method}} & \multirow{1}{*}{FID $\downarrow$}  & \multirow{1}{*}{IS $\uparrow$} &
\multirow{1}{*}{\textbf{Method}} & \multirow{1}{*}{FID $\downarrow$}
\\
\midrule
\multirow{2}{*}{Progressive-GAN \cite{karras2018progressive}}    &     \multirow{2}{*}{15.52}      &  \multirow{2}{*}{8.80 $\pm$ 0.05} & 
\multirow{2}{*}{SN-GAN \cite{miyato2018spectral}}    &     \multirow{2}{*}{40.1}      &  \multirow{2}{*}{9.16 $\pm$ 0.12}  &
\multirow{2}{*}{PAE \cite{bohm2020probabilistic}} &  \multirow{2}{*}{49.2} 
\\
\multirow{2}{*}{AutoGAN \cite{gong2019autogan}} & \multirow{2}{*}{12.42} &\multirow{2}{*}{8.55 $\pm$ 0.10} &
\multirow{2}{*}{Improving MMD-GAN \cite{wang2019improving}} & \multirow{2}{*}{37.64} &\multirow{2}{*}{9.23 $\pm$ 0.08}&
\multirow{2}{*}{BEGAN-CS \cite{chang2018escaping}} & \multirow{2}{*}{34.14}\\
\multirow{2}{*}{StyleGAN V2 \cite{karras2020analyzing}} & \multirow{2}{*}{11.07} &\multirow{2}{*}{9.18} &
\multirow{2}{*}{AutoGAN \cite{gong2019autogan}} & \multirow{2}{*}{31.01} &\multirow{2}{*}{9.16 $\pm$ 0.12} &
\multirow{2}{*}{PeerGAN \cite{wei2021peergan}} & \multirow{2}{*}{13.95}
\\
\multirow{2}{*}{Adversarial NAS-GAN \cite{gong2019autogan}} & \multirow{2}{*}{10.87} &\multirow{2}{*}{8.74 $\pm$ 0.07}&
\multirow{2}{*}{Adversarial NAS-GAN \cite{gao2020adversarialnas}} & \multirow{2}{*}{26.98} &\multirow{2}{*}{9.63 $\pm$ 0.19}&
\multirow{2}{*}{TransGAN-XL \cite{jiang2021transgan}}  & \multirow{2}{*}{12.23}
\\
\multirow{2}{*}{TransGAN-XL \cite{jiang2021transgan}} & \multirow{2}{*}{9.26} &\multirow{2}{*}{9.02 $\pm$ 0.11} &
\multirow{2}{*}{TransGAN-XL \cite{jiang2021transgan}} & \multirow{2}{*}{18.28} &\multirow{2}{*}{10.43 $\pm$ 0.17} &
\multirow{2}{*}{HDCGAN \cite{curto2020highresolution}} & \multirow{2}{*}{8.77}
\\
\multirow{2}{*}{StyleGAN2-ADA \cite{karras2020training}} & \multirow{2}{*}{2.92} &\multirow{2}{*}{9.83 $\pm$ 0.04}  &
\multirow{2}{*}{SNGAN-DCD  \cite{song2020discriminator}} & \multirow{2}{*}{17.68} &\multirow{2}{*}{9.33} &
\multirow{2}{*}{NCP-VAE \cite{aneja2020ncpvae}}  & \multirow{2}{*}{5.25}\\
\multirow{2}{*}{\textbf{Styleformer}}& \multirow{2}{*}{\textbf{2.82}}&
\multirow{2}{*}{\textbf{10.00 $\pm$ 0.12}} &
\multirow{2}{*}{\textbf{Styleformer}}& \multirow{2}{*}{\textbf{15.17}}&
\multirow{2}{*}{\textbf{11.01 $\pm$ 0.15}} &
\multirow{2}{*}{\textbf{Styleformer}}& \multirow{2}{*}{\textbf{3.92}}
\\
& & \\
\hline
\end{tabular}
}
\caption{Comparison results between Styleformer and other GAN models on low-resolution datasets. Results of other GAN models are collected from papers that reports their best results. We compute FID, IS in the same way as StyleGAN2-ADA, generating 50k images and compare their statistics against the 50k images from the training set for FID, computing the mean over 10 dependent trials using 5k generated images per trial for IS.}
\label{table:cifar-10}
\end{table*}

\begin{table}[!t]
\centering
\resizebox{1.0\linewidth}{!}{
\begin{tabular}{c|c|ccc}
\noalign{\smallskip}\noalign{\smallskip}\hline\hline
\multirow{2}{*}{\textbf{Dataset}} & \multirow{2}{*}{\textbf{Model}} & \multirow{2}{*}{FID $\downarrow$} & \multirow{2}{*}{Memory per GPU $\downarrow$} & \multirow{2}{*}{Speed $\downarrow$}\\
& & & &\\
\hline
\midrule
\multirow{2}{*}{CelebA}& Styleformer & 3.92 &14668MiB & 6.46 \\
& \textbf{Styleformer-L} & \textbf{3.36} & 5316MiB & 4.93\\
\midrule
\multirow{2}{*}{LSUN church}    
& Styleformer & -& OOM & -\\
& {\textbf{Styleformer-L}}  & {\textbf{7.99}} &8118MiB & 9.81\\ 
\midrule
\end{tabular}
}
\caption{Results on Styleformer-L which applies Linformer. ``Memory" is measured on 4 Titan-RTX with 16 batch size per GPU and ``Speed" means seconds for processing 1k images (sec/1kimg). We use the same hidden dimension and the number of layers in Styleformer and Styleformer-L.}
\label{table:linformer}
\end{table}

\section{Experiments}
\label{sec:4}
We only change the architecture of the generator in StyleGAN2-ADA, i.e., synthesis network, while maintaining the discriminator architecture and loss function. We use Fr$\Acute{e}$chet Inception Distance (FID) \cite{heusel2018gans} and Inception Score (IS) \cite{salimans2016improved}, evaluation metrics mainly used in the field of image generation. We compare our model with top GAN models such as StyleGAN2-ADA \cite{karras2020training}, and model related to our research such as TransGAN.
In Section \ref{section:4-1}, we show performance results of Styleformer in low-resolution dataset. Section \ref{section:4-2} provide evidence for a successful application of Linformer, including performance of Styleformer-L. In Section \ref{section:4-3}, we show high performance of Styleformer-C and prove the advantage and efficiency of our model by style mixing, and attention map visualization.

\begin{figure}[t]
\begin{center}
\includegraphics[height=.40\columnwidth]{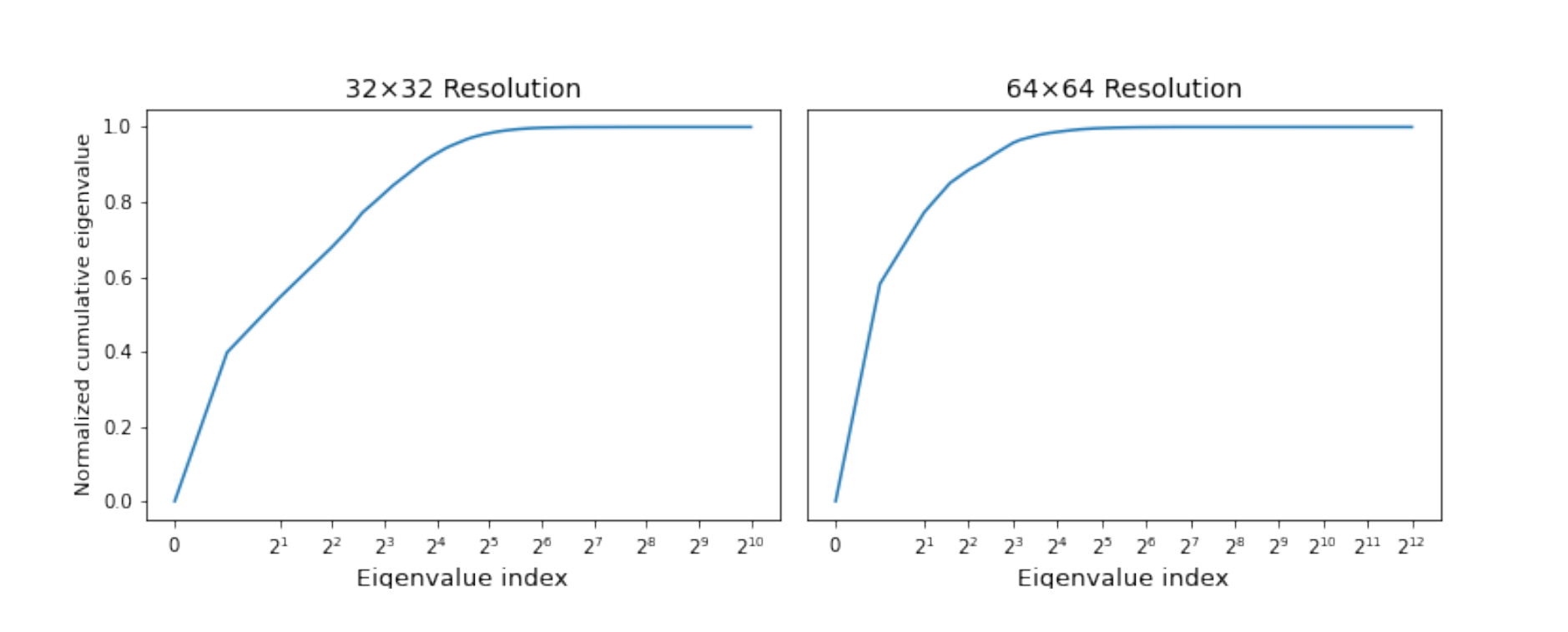}
\end{center}
\vspace{-5mm}
\caption{Spectrum analysis of attention map matrix at 32, 64 resolution.
We use pretrained Styleformer with CelebA dataset. 
}
\vspace{-3mm}
\label{fig:eigenvalue}
\end{figure}

\subsection{Low-Resolution Synthesis with Styleformer}
\label{section:4-1}
Styleformer achieves comparable performance to state-of-the-art in various low-resolution single-object datasets, including CIFAR-10 ($32 \times 32$) \cite{Krizhevsky09learningmultiple}, STL-10 ($48 \times 48$) \cite{pmlr-v15-coates11a}, and CelebA ($64 \times 64$) \cite{liu2015faceattributes}. 

As shown in Table \ref{table:cifar-10}, Styleformer outperforms prior GAN-based models, in terms of FID and IS. Especially in CIFAR-10, Styleformer records FID 2.82, and IS 10.00, which is comparable to current state-of-the-art and outperforming StyleGAN2-ADA-tuning. These results indicates that the Styleformer encoder has been modified to generate image successfully. Implementation details are in Appendix \ref{app:A}.

\begin{figure*}[t]
\begin{center}
\includegraphics[height=1\columnwidth, width=0.9\linewidth]{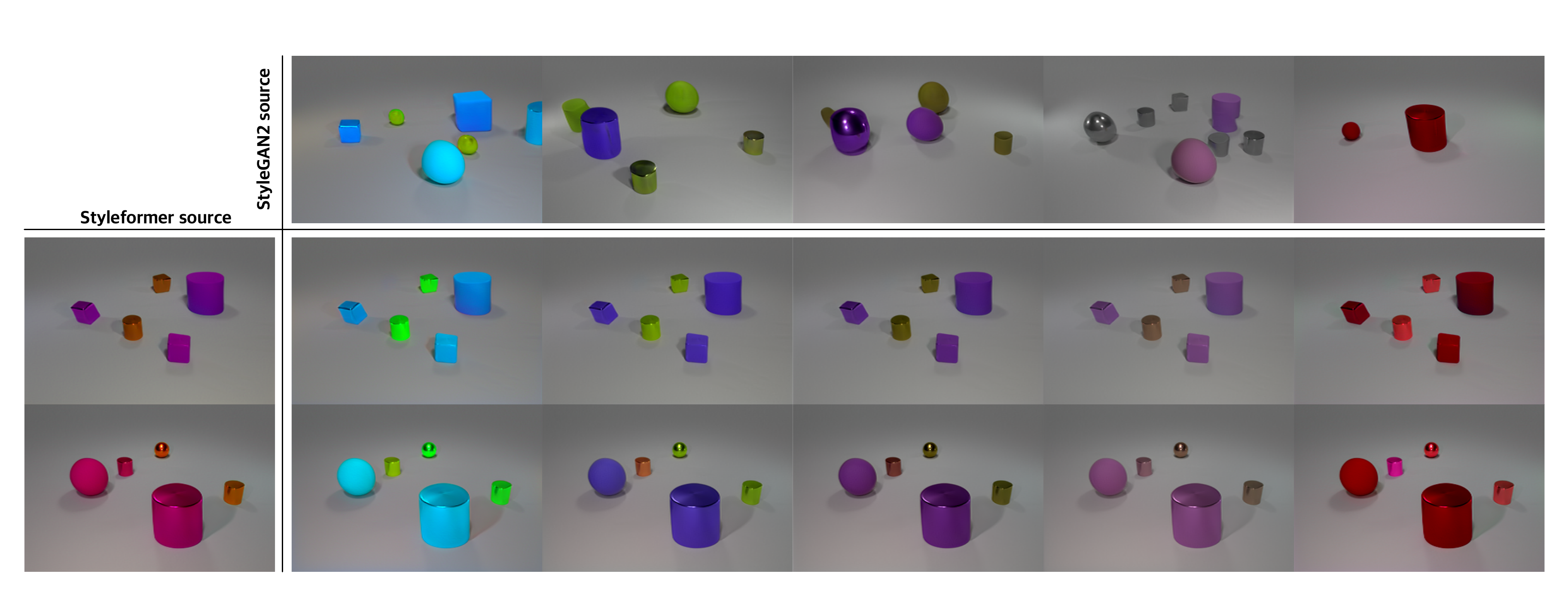}

\end{center}
\vspace{-5mm}
\caption{Style mixing experiment with Styleformer-C on CLEVR dataset. The images on the x-axis and y-axis were generated from their respective latent codes (StyleGAN2 source and Styleformer source, respectively); the rest of the images were generated by applying styles from Styleformer source to Styleformer at low resolution  and applying styles from StyleGAN2 source to StyleGAN2 at high resolution \cite{karras2019stylebased}. 
}
\vspace{-3mm}
\label{fig:style-mixing}
\end{figure*}

\begin{figure}[t]
\begin{center}
\includegraphics[height=.25\columnwidth, width=.30\columnwidth]{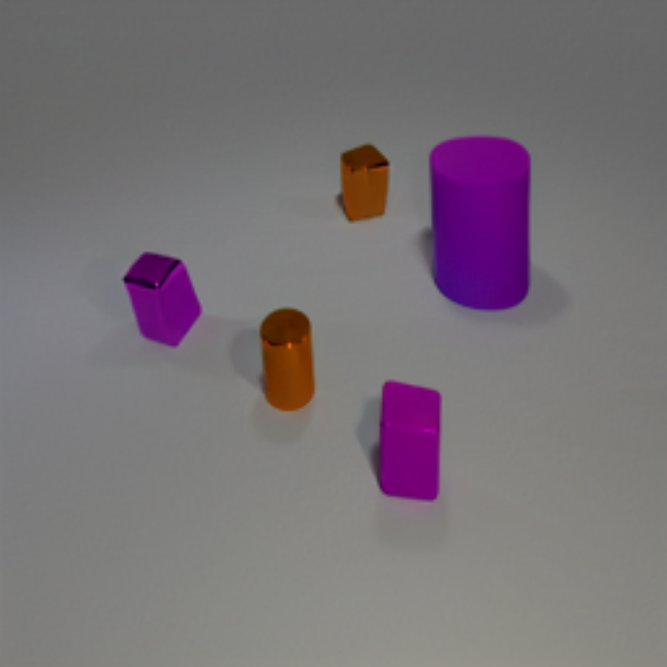}
\includegraphics[height=.25\columnwidth, width=.30\columnwidth]{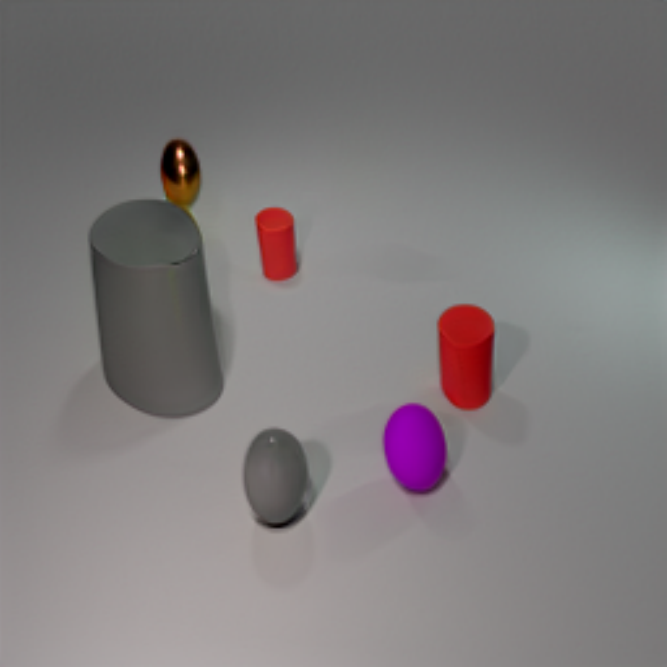}
\includegraphics[height=.25\columnwidth, width=.30\columnwidth]{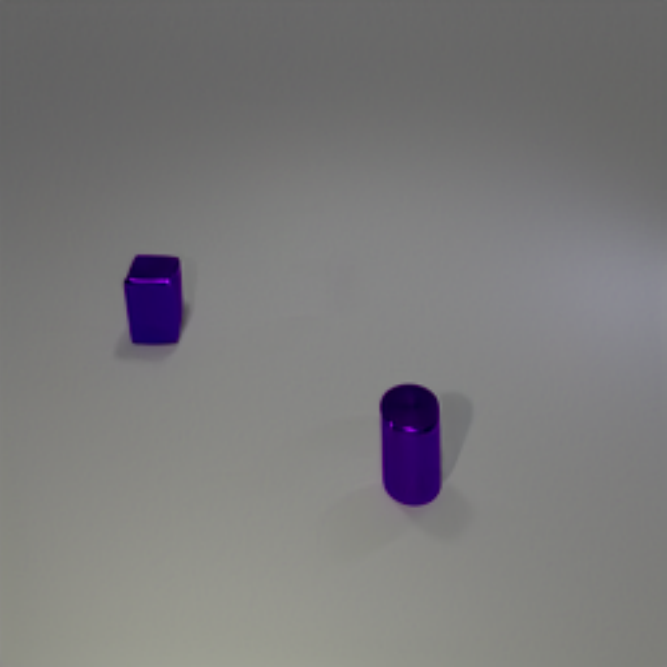}
\includegraphics[height=.25\columnwidth, width=.30\columnwidth]{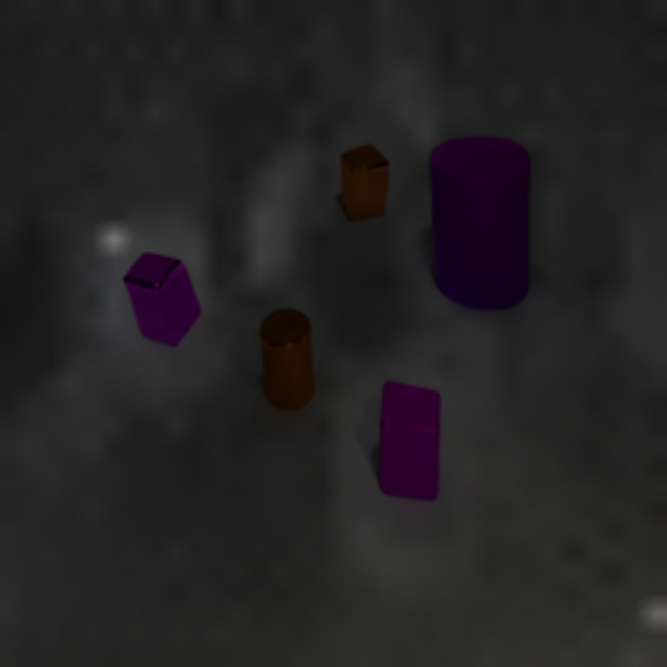}
\includegraphics[height=.25\columnwidth, width=.30\columnwidth]{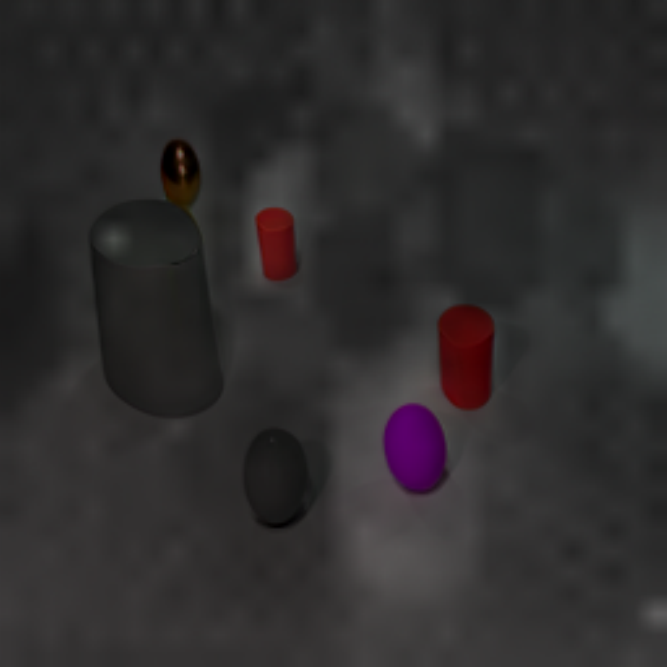}
\includegraphics[height=.25\columnwidth, width=.30\columnwidth]{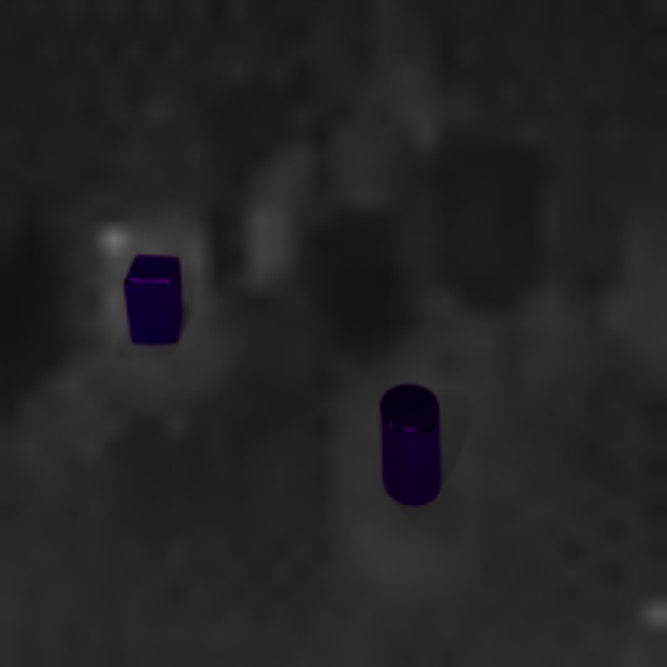}
\end{center}
\vspace{-5mm}
\caption{Visualizing attention map in generated CLEVR images.
}
\vspace{-3mm}
\label{fig:vis_attn}
\end{figure}

\subsection{Applying Linformer to Styleformer}
\label{section:4-2}
We experiment our method at Section \ref{section:3-4} which applies Linformer to Styleformer (Styleformer-L) on CelebA, $64 \times 64$ resolution, and LSUN-Church \cite{yu2016lsun} dataset resized to $128 \times 128$ resolution. 
As shown in Table \ref{table:linformer}, we find significant improvements in speed and memory and better performance than conventional Styleformer on CelebA. Memory performance is approximately three times more effective and speed performance is 1.3 times better in Styleformer-L. We also succeed in generating images of $128 \times 128$ resolution with the LSUN-Church dataset, which is difficult with pure Styleformer due to expensive memory. 

In addition, in the CelebA dataset, Styleformer-L shows higher performance in terms of FID than Styleformer, improving FID scores from 3.92 to 3.36. To analyze this phenomenon, we extract an attention map from Styleformer for generated CelebA images. As in \cite{wang2020linformer}, we apply singular value decomposition into attention map matrix, and plot the normalized cumulative singular value averaged over 1k generated images. As shown in Figure \ref{fig:eigenvalue}, most of the information in the attention map matrix can be recovered from the few large singular value, which means that the rank of attention map matrix is low. With low rank attention map, Linformer can be applied more efficiently \cite{wang2020linformer}.

Therefore, we show the possibility that when applying a self-attention operation for high-resolution images, it is not necessary to apply attention to all pixels and provide scalability to generate high-resolution images using Styleformer-L. See Appendix \ref{app:E} for implementation details.
    

\begin{table}[!t]
\centering
\resizebox{0.8\linewidth}{!}{
\begin{tabular}{c|c|c|c|c}
\noalign{\smallskip}\noalign{\smallskip}\hline\hline
\multirow{2}{*}{\textbf{Method}}
& \multicolumn{2}{c|}{\textbf{CLEVR}}& \multicolumn{2}{c}{\textbf{Cityscapes}}\\
& \multicolumn{1}{c}{FID $\downarrow$}& \multicolumn{1}{c}{IS $\uparrow$} & 
\multicolumn{1}{c}{FID $\downarrow$}& 
\multicolumn{1}{c}{IS $\uparrow$}\\
\midrule
\multirow{1}{*}{GAN \cite{goodfellow2014generative}}&     \multirow{1}{*}{25.02}&  \multirow{1}{*}{2.17} &     \multirow{1}{*}{11.57}&  \multirow{1}{*}{1.63}\\
\multirow{1}{*}{k-GAN \cite{sinha2020topk}}&     \multirow{1}{*}{28.09}&  \multirow{1}{*}{2.21} &     \multirow{1}{*}{51.08}&  \multirow{1}{*}{1.66}\\
\multirow{1}{*}{SAGAN \cite{zhang2019selfattention}}&     \multirow{1}{*}{26.04}&  \multirow{1}{*}{2.17} &     \multirow{1}{*}{12.81}&  \multirow{1}{*}{1.68}\\
\multirow{1}{*}{StyleGAN2 \cite{karras2020analyzing}}&     \multirow{1}{*}{16.05}&  \multirow{1}{*}{2.15} &     \multirow{1}{*}{8.35}&  \multirow{1}{*}{1.70}\\
\multirow{1}{*}{VQGAN \cite{esser2021taming}}&     \multirow{1}{*}{32.60}&  \multirow{1}{*}{2.03} &     \multirow{1}{*}{173.80}&  \multirow{1}{*}{\textbf{2.82}}\\
\multirow{1}{*}{Styleformer-C}&     \multirow{1}{*}{\textbf{11.67}}&  \multirow{1}{*}{\textbf{2.27}} &     \multirow{1}{*}{\textbf{5.99}}&  \multirow{1}{*}{2.56}\\
\midrule
\end{tabular}
}
\caption{Comparison between popular CNN based GAN models and Styleformer-C on CLEVR and Cityscapes. We use the results in \cite{hudson2021generative} for the performance of other models.}
\label{table:styleformer-C}
\end{table}

\subsection{Styleformer can Capture Global Interaction}
\label{section:4-3}
We experiment our method at Section \ref{section:3-4} which combines Styleformer and StyleGAN2 (Styleformer-C) on CLEVR($256 \times 256$) \cite{johnson2016clevr} and Cityscapes ($256 \times 256$) \cite{cordts2016cityscapes} for multi-object images and compositional scenes, AFHQ CAT ($512 \times 512$) \cite{choi2020stargan} for high-resolution single-object images. As shown in Table \ref{table:styleformer-C}, Styleformer-C records FID 11.67, IS 2.27 in CLEVR, and FID 5.99, IS 2.56 in Cityscapes which is comparable performance to current state-of-the-art, and showing better performance than StyleGAN2 in multi-object images and compositional scenes. This indirectly shows that Styleformer helps model to handle long-range dependency between components. 
 
To show more solid evidence that Styleformer captures global interaction, We conduct style mixing \cite{karras2019stylebased} in Styleformer-C. In detail, when generating new image from CLEVR dataset, we use two different latent codes $z_1$, $z_2$ and applying $z_1$ to Styleformer at low resolution and $z_2$ to StyleGAN2 at high resolution. As shown in Figure \ref{fig:style-mixing}, style corresponding to Styleformer (low-resolution) brings the basis for structural generation such as the location and structure of objects, while all colors or textures remain same. On the contrary, style corresponding to StyleGAN2 (high-resolution) brings the color and texture change, while maintaining location and shape of objects. This results directly prove that Styleformer controls global structure between objects, and handles long-range dependency. 

In addition, we visualize the attention map to provide more insight into the model's generating process. Figure \ref{fig:vis_attn} shows the concentration of attention to the position where the object exists. These visualizations show that the self-attention operation worked efficiently, enabling the model to perform long-range interaction, overcome the shortcoming of convolution operation.

\section{Conclusion}
\label{sec:con}
We propose Styleformer, a Transformer-based generative network that is novel and effective. We propose a method to efficiently generate images with self attention operation and achieve SOTA performance on various datasets. Furthermore, we propose Styleformer-L, which reduces the complex computation to linear, enabling to generate high-resolution images. We also present a method of efficiently generating a compositional scene while capturing with long-range dependency through Styleformer-C. 
There still seems to be room for improvement, such as reducing computation cost, but we hope that our work will speed up the application of Transformers to the field of computer vision, helping the development of the computer vision field. However, development of the generative model can create fake media data using synthesized face images (e.g. deepfake), so particular attention should be paid in the future.


{\small
\bibliographystyle{ieee_fullname}
\bibliography{egbib}

\begin{thebibliography}{10}\itemsep=-1pt

\bibitem{aneja2020ncpvae}
Jyoti Aneja, Alexander Schwing, Jan Kautz, and Arash Vahdat.
\newblock Ncp-vae: Variational autoencoders with noise contrastive priors,
  2020.

\bibitem{arjovsky2017wasserstein}
Martin Arjovsky, Soumith Chintala, and Léon Bottou.
\newblock Wasserstein gan, 2017.

\bibitem{belousov2021mobilestylegan}
Sergei Belousov.
\newblock Mobilestylegan: A lightweight convolutional neural network for
  high-fidelity image synthesis, 2021.

\bibitem{bertasius2021spacetime}
Gedas Bertasius, Heng Wang, and Lorenzo Torresani.
\newblock Is space-time attention all you need for video understanding?, 2021.

\bibitem{bohm2020probabilistic}
Vanessa Böhm and Uroš Seljak.
\newblock Probabilistic auto-encoder, 2020.

\bibitem{chang2018escaping}
Chia-Che Chang, Chieh~Hubert Lin, Che-Rung Lee, Da-Cheng Juan, Wei Wei, and
  Hwann-Tzong Chen.
\newblock Escaping from collapsing modes in a constrained space, 2018.

\bibitem{choi2018stargan}
Yunjey Choi, Minje Choi, Munyoung Kim, Jung-Woo Ha, Sunghun Kim, and Jaegul
  Choo.
\newblock Stargan: Unified generative adversarial networks for multi-domain
  image-to-image translation, 2018.

\bibitem{choi2020stargan}
Yunjey Choi, Youngjung Uh, Jaejun Yoo, and Jung-Woo Ha.
\newblock Stargan v2: Diverse image synthesis for multiple domains, 2020.

\bibitem{pmlr-v15-coates11a}
Adam Coates, Andrew Ng, and Honglak Lee.
\newblock An analysis of single-layer networks in unsupervised feature
  learning.
\newblock In Geoffrey Gordon, David Dunson, and Miroslav Dudík, editors, {\em
  Proceedings of the Fourteenth International Conference on Artificial
  Intelligence and Statistics}, volume~15 of {\em Proceedings of Machine
  Learning Research}, pages 215--223, Fort Lauderdale, FL, USA, 11--13 Apr
  2011. PMLR.

\bibitem{cordts2016cityscapes}
Marius Cordts, Mohamed Omran, Sebastian Ramos, Timo Rehfeld, Markus Enzweiler,
  Rodrigo Benenson, Uwe Franke, Stefan Roth, and Bernt Schiele.
\newblock The cityscapes dataset for semantic urban scene understanding, 2016.

\bibitem{curto2020highresolution}
J.~D. Curtó, I.~C. Zarza, Fernando de~la Torre, Irwin King, and Michael~R.
  Lyu.
\newblock High-resolution deep convolutional generative adversarial networks,
  2020.

\bibitem{dosovitskiy2020image}
Alexey Dosovitskiy, Lucas Beyer, Alexander Kolesnikov, Dirk Weissenborn,
  Xiaohua Zhai, Thomas Unterthiner, Mostafa Dehghani, Matthias Minderer, Georg
  Heigold, Sylvain Gelly, Jakob Uszkoreit, and Neil Houlsby.
\newblock An image is worth 16x16 words: Transformers for image recognition at
  scale, 2020.

\bibitem{dumoulin2018feature-wise}
Vincent Dumoulin, Ethan Perez, Nathan Schucher, Florian Strub, Harm~de Vries,
  Aaron Courville, and Yoshua Bengio.
\newblock Feature-wise transformations.
\newblock {\em Distill}, 2018.
\newblock https://distill.pub/2018/feature-wise-transformations.

\bibitem{dumoulin2017learned}
Vincent Dumoulin, Jonathon Shlens, and Manjunath Kudlur.
\newblock A learned representation for artistic style, 2017.

\bibitem{esser2021taming}
Patrick Esser, Robin Rombach, and Björn Ommer.
\newblock Taming transformers for high-resolution image synthesis, 2021.

\bibitem{gabbay2019style}
Aviv Gabbay and Yedid Hoshen.
\newblock Style generator inversion for image enhancement and animation, 2019.

\bibitem{gao2020adversarialnas}
Chen Gao, Yunpeng Chen, Si Liu, Zhenxiong Tan, and Shuicheng Yan.
\newblock Adversarialnas: Adversarial neural architecture search for gans,
  2020.

\bibitem{ghiasi2017exploring}
Golnaz Ghiasi, Honglak Lee, Manjunath Kudlur, Vincent Dumoulin, and Jonathon
  Shlens.
\newblock Exploring the structure of a real-time, arbitrary neural artistic
  stylization network, 2017.

\bibitem{pmlr-v9-glorot10a}
Xavier Glorot and Yoshua Bengio.
\newblock Understanding the difficulty of training deep feedforward neural
  networks.
\newblock In Yee~Whye Teh and Mike Titterington, editors, {\em Proceedings of
  the Thirteenth International Conference on Artificial Intelligence and
  Statistics}, volume~9 of {\em Proceedings of Machine Learning Research},
  pages 249--256, Chia Laguna Resort, Sardinia, Italy, 13--15 May 2010. PMLR.

\bibitem{gong2019autogan}
Xinyu Gong, Shiyu Chang, Yifan Jiang, and Zhangyang Wang.
\newblock Autogan: Neural architecture search for generative adversarial
  networks, 2019.

\bibitem{goodfellow2014generative}
Ian~J. Goodfellow, Jean Pouget-Abadie, Mehdi Mirza, Bing Xu, David
  Warde-Farley, Sherjil Ozair, Aaron Courville, and Yoshua Bengio.
\newblock Generative adversarial networks, 2014.

\bibitem{graham2021levit}
Ben Graham, Alaaeldin El-Nouby, Hugo Touvron, Pierre Stock, Armand Joulin,
  Hervé Jégou, and Matthijs Douze.
\newblock Levit: a vision transformer in convnet's clothing for faster
  inference, 2021.

\bibitem{heusel2018gans}
Martin Heusel, Hubert Ramsauer, Thomas Unterthiner, Bernhard Nessler, and Sepp
  Hochreiter.
\newblock Gans trained by a two time-scale update rule converge to a local nash
  equilibrium, 2018.

\bibitem{howard2017mobilenets}
Andrew~G. Howard, Menglong Zhu, Bo Chen, Dmitry Kalenichenko, Weijun Wang,
  Tobias Weyand, Marco Andreetto, and Hartwig Adam.
\newblock Mobilenets: Efficient convolutional neural networks for mobile vision
  applications, 2017.

\bibitem{huang2017arbitrary}
Xun Huang and Serge Belongie.
\newblock Arbitrary style transfer in real-time with adaptive instance
  normalization, 2017.

\bibitem{hudson2021generative}
Drew~A. Hudson and C.~Lawrence Zitnick.
\newblock Generative adversarial transformers, 2021.

\bibitem{isola2018imagetoimage}
Phillip Isola, Jun-Yan Zhu, Tinghui Zhou, and Alexei~A. Efros.
\newblock Image-to-image translation with conditional adversarial networks,
  2018.

\bibitem{jiang2021transgan}
Yifan Jiang, Shiyu Chang, and Zhangyang Wang.
\newblock Transgan: Two transformers can make one strong gan, 2021.

\bibitem{johnson2018image}
Justin Johnson, Agrim Gupta, and Li Fei-Fei.
\newblock Image generation from scene graphs, 2018.

\bibitem{johnson2016clevr}
Justin Johnson, Bharath Hariharan, Laurens van~der Maaten, Li Fei-Fei,
  C.~Lawrence Zitnick, and Ross Girshick.
\newblock Clevr: A diagnostic dataset for compositional language and elementary
  visual reasoning, 2016.

\bibitem{karras2018progressive}
Tero Karras, Timo Aila, Samuli Laine, and Jaakko Lehtinen.
\newblock Progressive growing of gans for improved quality, stability, and
  variation, 2018.

\bibitem{karras2020training}
Tero Karras, Miika Aittala, Janne Hellsten, Samuli Laine, Jaakko Lehtinen, and
  Timo Aila.
\newblock Training generative adversarial networks with limited data, 2020.

\bibitem{karras2019stylebased}
Tero Karras, Samuli Laine, and Timo Aila.
\newblock A style-based generator architecture for generative adversarial
  networks, 2019.

\bibitem{karras2020analyzing}
Tero Karras, Samuli Laine, Miika Aittala, Janne Hellsten, Jaakko Lehtinen, and
  Timo Aila.
\newblock Analyzing and improving the image quality of stylegan, 2020.

\bibitem{kolesnikov2020big}
Alexander Kolesnikov, Lucas Beyer, Xiaohua Zhai, Joan Puigcerver, Jessica Yung,
  Sylvain Gelly, and Neil Houlsby.
\newblock Big transfer (bit): General visual representation learning, 2020.

\bibitem{Krizhevsky09learningmultiple}
Alex Krizhevsky.
\newblock Learning multiple layers of features from tiny images.
\newblock Technical report, 2009.

\bibitem{NIPS2012_c399862d}
Alex Krizhevsky, Ilya Sutskever, and Geoffrey~E Hinton.
\newblock Imagenet classification with deep convolutional neural networks.
\newblock In F. Pereira, C.~J.~C. Burges, L. Bottou, and K.~Q. Weinberger,
  editors, {\em Advances in Neural Information Processing Systems}, volume~25.
  Curran Associates, Inc., 2012.

\bibitem{ledig2017photorealistic}
Christian Ledig, Lucas Theis, Ferenc Huszar, Jose Caballero, Andrew Cunningham,
  Alejandro Acosta, Andrew Aitken, Alykhan Tejani, Johannes Totz, Zehan Wang,
  and Wenzhe Shi.
\newblock Photo-realistic single image super-resolution using a generative
  adversarial network, 2017.

\bibitem{liu2021swin}
Ze Liu, Yutong Lin, Yue Cao, Han Hu, Yixuan Wei, Zheng Zhang, Stephen Lin, and
  Baining Guo.
\newblock Swin transformer: Hierarchical vision transformer using shifted
  windows, 2021.

\bibitem{liu2015faceattributes}
Ziwei Liu, Ping Luo, Xiaogang Wang, and Xiaoou Tang.
\newblock Deep learning face attributes in the wild.
\newblock In {\em Proceedings of International Conference on Computer Vision
  (ICCV)}, December 2015.

\bibitem{mao2017squares}
Xudong Mao, Qing Li, Haoran Xie, Raymond Y.~K. Lau, Zhen Wang, and Stephen~Paul
  Smolley.
\newblock Least squares generative adversarial networks, 2017.

\bibitem{miyato2018spectral}
Takeru Miyato, Toshiki Kataoka, Masanori Koyama, and Yuichi Yoshida.
\newblock Spectral normalization for generative adversarial networks, 2018.

\bibitem{radford2016unsupervised}
Alec Radford, Luke Metz, and Soumith Chintala.
\newblock Unsupervised representation learning with deep convolutional
  generative adversarial networks, 2016.

\bibitem{salimans2016improved}
Tim Salimans, Ian Goodfellow, Wojciech Zaremba, Vicki Cheung, Alec Radford, and
  Xi Chen.
\newblock Improved techniques for training gans, 2016.

\bibitem{sinha2020topk}
Samarth Sinha, Zhengli Zhao, Anirudh Goyal, Colin Raffel, and Augustus Odena.
\newblock Top-k training of gans: Improving gan performance by throwing away
  bad samples, 2020.

\bibitem{song2020discriminator}
Yuxuan Song, Qiwei Ye, Minkai Xu, and Tie-Yan Liu.
\newblock Discriminator contrastive divergence: Semi-amortized generative
  modeling by exploring energy of the discriminator, 2020.

\bibitem{tan2020efficientnet}
Mingxing Tan and Quoc~V. Le.
\newblock Efficientnet: Rethinking model scaling for convolutional neural
  networks, 2020.

\bibitem{tolstikhin2021mlpmixer}
Ilya Tolstikhin, Neil Houlsby, Alexander Kolesnikov, Lucas Beyer, Xiaohua Zhai,
  Thomas Unterthiner, Jessica Yung, Andreas Steiner, Daniel Keysers, Jakob
  Uszkoreit, Mario Lucic, and Alexey Dosovitskiy.
\newblock Mlp-mixer: An all-mlp architecture for vision, 2021.

\bibitem{vaswani2017attention}
Ashish Vaswani, Noam Shazeer, Niki Parmar, Jakob Uszkoreit, Llion Jones,
  Aidan~N. Gomez, Lukasz Kaiser, and Illia Polosukhin.
\newblock Attention is all you need, 2017.

\bibitem{wang2020linformer}
Sinong Wang, Belinda~Z. Li, Madian Khabsa, Han Fang, and Hao Ma.
\newblock Linformer: Self-attention with linear complexity, 2020.

\bibitem{wang2019improving}
Wei Wang, Yuan Sun, and Saman Halgamuge.
\newblock Improving mmd-gan training with repulsive loss function, 2019.

\bibitem{wei2021peergan}
Jiaheng Wei, Minghao Liu, Jiahao Luo, Qiutong Li, James Davis, and Yang Liu.
\newblock Peergan: Generative adversarial networks with a competing peer
  discriminator, 2021.

\bibitem{wu2021cvt}
Haiping Wu, Bin Xiao, Noel Codella, Mengchen Liu, Xiyang Dai, Lu Yuan, and Lei
  Zhang.
\newblock Cvt: Introducing convolutions to vision transformers, 2021.

\bibitem{yu2016lsun}
Fisher Yu, Ari Seff, Yinda Zhang, Shuran Song, Thomas Funkhouser, and Jianxiong
  Xiao.
\newblock Lsun: Construction of a large-scale image dataset using deep learning
  with humans in the loop, 2016.

\bibitem{yu2018generative}
Jiahui Yu, Zhe Lin, Jimei Yang, Xiaohui Shen, Xin Lu, and Thomas~S. Huang.
\newblock Generative image inpainting with contextual attention, 2018.

\bibitem{zhang2019selfattention}
Han Zhang, Ian Goodfellow, Dimitris Metaxas, and Augustus Odena.
\newblock Self-attention generative adversarial networks, 2019.

\bibitem{zheng2021rethinking}
Sixiao Zheng, Jiachen Lu, Hengshuang Zhao, Xiatian Zhu, Zekun Luo, Yabiao Wang,
  Yanwei Fu, Jianfeng Feng, Tao Xiang, Philip H.~S. Torr, and Li Zhang.
\newblock Rethinking semantic segmentation from a sequence-to-sequence
  perspective with transformers, 2021.

\bibitem{zhu2020unpaired}
Jun-Yan Zhu, Taesung Park, Phillip Isola, and Alexei~A. Efros.
\newblock Unpaired image-to-image translation using cycle-consistent
  adversarial networks, 2020.

\bibitem{Zhu_2020}
Peihao Zhu, Rameen Abdal, Yipeng Qin, and Peter Wonka.
\newblock Sean: Image synthesis with semantic region-adaptive normalization.
\newblock {\em 2020 IEEE/CVF Conference on Computer Vision and Pattern
  Recognition (CVPR)}, Jun 2020.

\bibitem{zhu2020semantically}
Zhen Zhu, Zhiliang Xu, Ansheng You, and Xiang Bai.
\newblock Semantically multi-modal image synthesis, 2020.

\end{thebibliography}
}

\newpage

\appendix


\section{Implementation Details of Styleformer}
\label{app:A}
We implemented our Styleformer on top of the StyleGAN2-ADA Pytorch implementation \footnote{\url{https://github.com/NVlabs/stylegan2-ada-pytorch}}. 
Most of the details have not changed except generator architecture. We used CIFAR-10 tuning version of StyleGAN2-ADA, which means disabling style mixing regularization, path length regularization, and residual connections in D when training. We also fixed mapping network's depth to 2. We used bilinear filtering in all upsampling layer used in Styleformer. We use data augmentation pipeline suggested in StyleGAN2-ADA, and did not use mixed-precision training for all experiments.

For Styleformer encoder, we add bias and noise at the end of the encoder block, then performing leaky RELU with $\alpha=0.2$. After passing several encoder blocks, we reshaped it to the form of square feature map, i.e., Unflatten. We then proceed bilinear upsample operation as we said in the Section \ref{section:3-1}, but also convert these reshaped output for each resolution into an RGB channel, using ToRGB layer. We upsample each of RGB output and add to each other, creating an output-skip connection generator, similar to StyleGAN2 generator. Originally ToRGB layer converts high-dimensional per pixel data into RGB per pixel data via $1 \times 1$ convolution operation, which we replace it to same operation, linear operation. We initialize all weights in Styleformer encoder using same method used in Pytorch linear layer. Unlike StyleGAN2-ADA, we employ weight demodulation also in ToRGB layer. We perform all experiments with 4 Titan-RTX using Pytorch 1.7.1. All of our experiments presented in the paper including failure spent about four months.

As in Section \ref{section:4-1}, the number of the Styleformer encoder and hidden dimension size for each resolution can be chosen as hyperparameters. We call these two hyperparameters as "Layers" and "Hidden size", respectively. 

\paragraph{Low-Resolution Synthesis with Styleformer}
For low-resolution synthesis experiment in Section \ref{section:4-1}, we use pure Styleformer. Each Layers and Hidden size used for CIFAR-10, STL-10, and CelebA are shown in Table \ref{table:size model}. We trained Styleformer for 65M, 92M, 25M at CIFAR-10, STL-10 and CelebA, respectively.

For CIFAR-10 experiment, we use 50K images ($32 \times 32$) at the training set, without using the label. For STL-10 experiment, we resize $96 \times 96$ image datasets to $48 \times 48$, and using 5k training images, 100k unlabeled images together as in \cite{jiang2021transgan}. We change the size of the constant input from $8 \times 8$ to $12 \times 12$ to generate an image with a size of $48 \times 48$. For CelebA dataset, we use 200k unlabeled face images of the Align and Cropped version, which we resize to $64 \times 64$ resolution as in \cite{jiang2021transgan}. We start at $8 \times 8$ constant, as we train CIFAR-10 dataset.

\begin{table}[!t]
\centering
\caption{Details of Styleformer hyperparameters at low resolution synthesis. This model setting match with performance result at Table 2 in paper.}

\label{table:size model}
\resizebox{0.9\linewidth}{!}{
\begin{tabular}{c|ccc}
\noalign{\smallskip}\noalign{\smallskip}\hline\hline
\multirow{2}{*}{\textbf{Datasets}}& \multirow{2}{*}{Layers} & \multirow{2}{*}{Hidden size} & \multirow{2}{*}{FID} \\
& & & \\
\midrule
CIFAR-10 & \{1,3,3\} & \{1024,512,512\} & 2.82\\
STL-10 & \{1,2,2\} & \{1024,256,64\} & 15.17\\
CelebA & \{1,2,1,1\} & \{1024,256,64,64\} & 3.92\\
\midrule
\end{tabular}
}
\end{table}

\paragraph{Ablation study details}
As we said in paper, we use small version of Styleformer with CIFAR-10 dataset for ablation study. In more detail, we use {1,2,2} for Layers, {256, 64, 16} for Hidden size, trained for 20M images. 

\paragraph{Number of head experiment}
We conduct experiment about number of heads effect using one layer Styleformer, as said in Figure \ref{fig:multihead_graph}. One Layer Styleformer is a model that starts with $32 \times 32$ learned constant and only have one Styleformer encoder with hidden dimension size 256. We trained the model for 20M images, same as ablation study.

\section{Attention map analysis}
\label{app:B}
\begin{figure*}[t]
\begin{center}
\includegraphics[width=1.8\columnwidth]{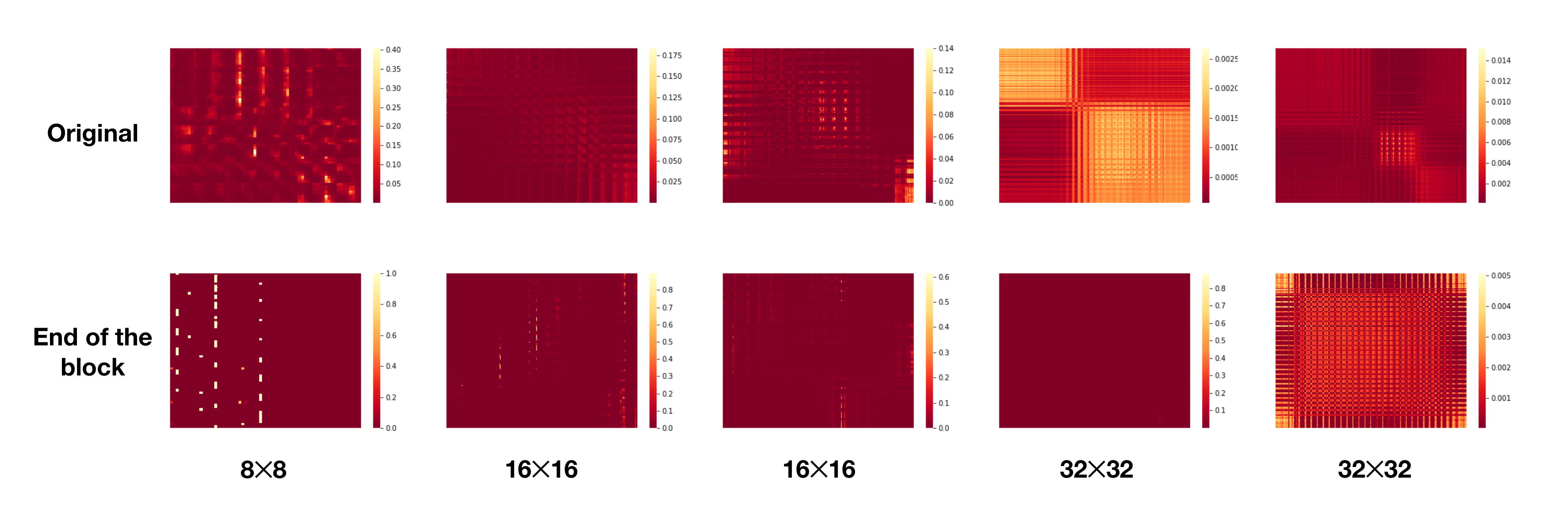}
\end{center}
\vspace{-5mm}
\caption{
Comparison of attention map experimented based on Layer Normalization
}
\vspace{-3mm}
\label{fig:layer_attn}
\end{figure*}

\paragraph{Post-Layer normalization}
We analyze the results of the attention map experiment on layer normalization. We propose in Section \ref{section:3-2} that if we perform layer normalization at the end of Styleformer encoder, it breaks the attention map. Figure \ref{fig:layer_attn} shows that if layer normalization is located at the end of the encoder, the attention map has not been properly learned. Therefore, we position layer normalization to the front of the encoder (i.e. after applying style modulation) so that the attention map can effectively learn the relationship between pixels. The experiment is all conducted on CIFAR-10, using small version of Styleformer, same as ablation study. 

\begin{figure*}[t]
\begin{center}
\includegraphics[width=1.82\columnwidth]{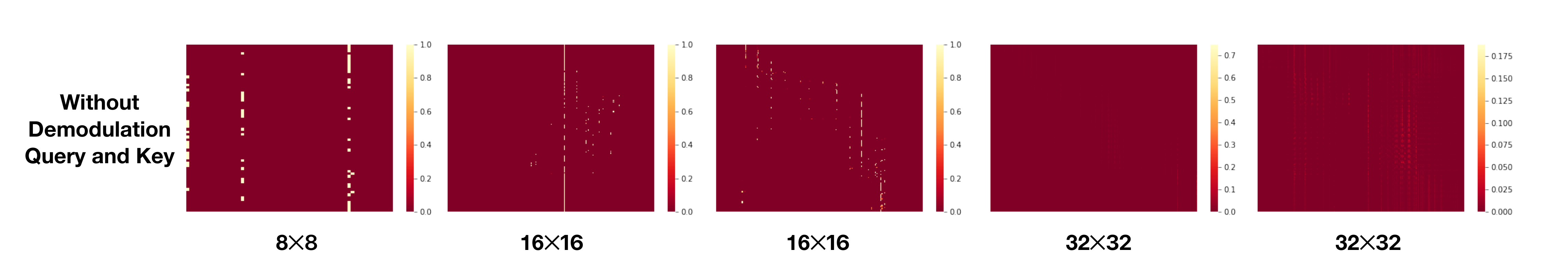}
\end{center}
\vspace{-5mm}
\caption{
Attention map without demodulation operation to query and key. Comparing with Figure \ref{fig:layer_attn} upper row, we can see specific large value in attention map.
}
\vspace{-3mm}
\label{fig:demod_attn}
\end{figure*}

\paragraph{Demodulation for Query and Key}
As in Section \ref{section:3-3}, we demonstrate that when an attention map is created with $Q$, $K$ from input scaled by style vector, specific value in attention map becomes very large. We show the attention map without demodulation operation to $Q$ and 
$K$ in Styleformer at Figure \ref{fig:layer_attn}. We can see that without demodulation operation, attention is heavily concentrated on a particular pixel, preventing the attention operation from working properly. We overcome this problem with demodulation operation to $Q$, $K$, before creating attention map.

\section{Intuition of Increased Multi-Head Attention}
\label{app:C}
Transformer is designed for natural language processing (NLP). It is difficult to use original 
Transformer in the field of image generation.
A convolution network can efficiently generate images, unlike a linear layer, because it proceeds both operations between pixels using kernels and between channels. Although not a convolution network structure, \cite{tolstikhin2021mlpmixer} has shown that architecture with pixel-to-pixel and channel-to-channel operations can be efficient in learning information about images.

Unlike the original convolution operation, in MobileStyleGAN \cite{belousov2021mobilestylegan}, they show good performance in image generation even by separating the interpixel and interchannel operations with depthwise separable convolution.

Main operation of Transformer can be divided into a interpixel operation (i.e. pixel section) and a interchannel operation (i.e. channel section) and it can be expressed by the following formula.
\begin{align}
    \label{eq:2}
    A_{i} = soft&max(\frac{Q_{i}K_{i}^{T}}{\sqrt{d_{k}}}),\\
    \label{eq:3}
    head_{i} &= A_{i}V_{i},\\
    \label{eq:4}
    Multihead(Q, K, V) = &Concat(head_{1},\dots,head_{k})W^{O},
\end{align}
In above formulation, $Q_{i}$, $K_{i}$, $V_{i}$ is query, key, value for each head. $d$ is the dimension of ($query, key, value$), $k$ is the number of heads, and $d_{k}$ is $d / k$.  $A_{i}$ is attention map and $W^{O}$ is the parameters of a linear layer that integrates multi-heads.  

 Pixel section corresponds for the self-attention operation between pixels (i.e., Equation \ref{eq:3}), and channel section corresponds for integration of multi-head with linear layer, which operates between channels (i.e., Equation \ref{eq:4}). In a Transformer, the pixel section is slightly different from depthwise convolution. In depthwise convolution, kernel weights exist for each channel, but in Transformer, attention map $A$ acts like one huge kernel, which means applying equal kernel weight to all different channels in $V$.  
It is difficult to create a powerful generator using a Transformer because the same attention kernel is applied for each channel, unlike the generator using depthwise separable convolution. 

Using increased multi-head attention, this problem could be overcome. We can generate various attention maps (i.e., kernels) by increasing the number of heads. However, increasing the number of heads inevitably leads to smaller depth, where depth is hidden channel dimension divided by the number of heads. Since depth is a dimension used to create the attention map, there exists a minimum depth required. However, due to differences in the properties of pixels and tokens, the required depth size in computer vision is smaller than NLP.

In the field of NLP, a single token has a lot of information, so the required depth dimension, which represents the token, must be large, but one pixel in an image has less information than a token, which means that the required depth is smaller than a traditional Transformer. 
As described in the caption of Figure \ref{fig:multihead_graph}, the left graph shows a hidden dimension of 256, and the right graph shows 32. In the left graph, until the number of heads is 8 (depth is 32), performance increases when the number of heads grows up, but after that, even if the number of heads is increased, the performance decreases because the depth is less than 32. Similarly, in the right graph, performance is best when the number of heads is 1 (depth is 32), and after that, performance degrades because the depth is less than 32.  From these results, we demonstrate that increasing the number of heads too much results in poor performance, and fixing the depth to 32.

Meanwhile, the channel section is a layer to integrate the multi-head together, where the linear layer is exactly the same operation as the $1\times1$ convolution, i.e., pointwise convolution.
Unlike convolution, the attention kernel is a kernel generated by the input itself, so it can create a more dense kernel, and it is advantageous to capture global features because it considers the relationship between all pixels.
Therefore, by enhancing multi-head attention, the Styleformer can play a more powerful role as depthwise separable convolution, enabling generate high-quality images.

\section{Demodulation for Encoder Output}
\label{app:D}
As described in Section \ref{section:3-3}, self-attention conducts more operations than the convolution operation, making the demodulation process more complex. We show how standard deviation of encoder output can be derived (Demodulation for Encoder Output at Section \ref{section:3-3}), and the effect of the number of pixel in output standard deviation.

\paragraph{Derivation of encoder output standard deviation}
After demodulation operation to $Q$, $K$ and $V$, Styleformer encoder performs style modulation to input $V$, weighted sum of $V$ with attention map, and then performs linear operation. Let's consider the attention map matrix as $A$, and the weight matrix of linear operation as $w$. Since the matrix multiplication is associative, statistics of the output are the same even if the linear operation is calculated before multiplication with attention map :
\begin{align}
    \label{eq:99}
    output = [A(s_j \cdot V)] w = A [(s_j \cdot V)w].
\end{align}
From now on, we think the linear operation is conducted before the multiplication with the attention map. Therefore, same as demodulation for query, key, and value, style modulation to $V$ can be replaced to scaling linear weights:
\begin{align}
    \label{eq:100}
    w'_{jk} = s_j \cdot w_{jk},
\end{align}
where $s_{j}$ scales $j$th feature map of $V$, and $k$ enumerates the flattened output feature map. Assuming that input $V$ have a unit standard deviation, the standard deviation of output after linear operation can be derived as follows:
\begin{align}
    \label{eq:101}
    \sigma_{k} &= \sqrt{\sum_{j}{w^{'}_{jk}{^{2}}}}.
\end{align}
Finally, this output is multiplied with the attention map matrix $A$. When the attention score vector for $l$th pixel is expressed as $A_{l \cdot}$, standard deviation of encoder output is as follows:
\begin{align}
    \label{eq:103}
    \sigma^{'}_{lk} &= \sqrt{\sum_{\cdot}{A_{l\cdot}{^2}} \cdot \sum_{j}{w^{'}_{jk}{^2}}}.
\end{align}

\paragraph{Effect of the number of pixel}
Scaling the encoder output feature map $k$ with  $1/\sigma^{''}_{k}$ where
    $\sigma^{''}_{k} = \sqrt{\sum_{j}{w^{'}_{jk}{^2}}}$, the standard deviation of output activations will be 
\begin{align}
    \sigma_{lk} = \sqrt{\sum_{\cdot}{A_{l\cdot}{^2}}}.
\end{align}
Since the attention map $A$ is matrix after softmax operation, $\sum_{\cdot}{A_{l\cdot}}=1$. Assuming $A_{l\cdot}$ are i.i.d random variables from normal distribution with mean $\cfrac{1}{n}$ and variance $\cfrac{1}{n{^2}}$, and $n$ denotes the number of pixel, $\sigma_{lk}{^2}$ can be derived as follows:
\begin{align}
    \sigma_{lk}{^2} = \sum_{\cdot}{A_{l\cdot}{^2}} = \sum_{\cdot}{(A_{l\cdot}-\cfrac{1}{n})}{^2} + \cfrac{1}{n},
\end{align}
using $\sum_{\cdot}{A_{l\cdot}}=1$. Then $(A_{l\cdot}-\cfrac{1}{n})$ become random variables from normal distribution with zero mean and variance $1/n{^2}$. Based on the property of normal distribution, $\sum_{\cdot}{(A_{l\cdot}-\cfrac{1}{n})}$ follows gamma distribution with shape parameter $\cfrac{n}{2}$ and scale parameter $\cfrac{2}{n{^2}}$. Therefore, using Chebyshev inequality, we have 
\begin{align}
    \Pr(|\sum_{\cdot}{(A_{l\cdot}-\cfrac{1}{n})}{^2}-\cfrac{1}{n}|\leq\cfrac{1}{n})\geq1-\cfrac{2}{n},
\end{align}
meaning $\sigma_{lk}$ approaches zero when the number of pixels(i.e., $n$) increase. 

\section{Implementation details of Styleformer-L}
\label{app:E}
We introduce Styleformer-L, a model with Linformer applied to self-attention operation of Styleformer. When applying Linformer, we fix $k$ (i.e., projection dimension for key, value) to 256, and apply to the encoder block above $32 \times 32$ resolution. For projection parameter sharing effect, we used Key-value sharing. It means we create single $E$ projection matrix for each layer that applies equally to the key, value of each head. We project key and value to k dimension after demodulation for key and value, i.e., before creating attention map and weight sum of value. When using Linformer, to prevent augmentation leaking, we clamp augmentation probability to 0.7.

\paragraph{CelebA}
We conduct CelebA experiment using Styleformer-L in the same setting as CelebA using pure Styleformer for fair comparision (Table 3 in paper). We use $\{1,2,1,1\}$ for Layers, $\{1024,256,64,64\}$ for Hidden size, and training for 25M images.

\paragraph{LSUN-Church} 
We use $\{1,2,1,1,1\}$ for Layers, $\{1024,256,64,64,64\}$ for Hidden size, training for 40M images.

\section{Implementation Details of Styleformer-C}
\label{app:F}
We introduce Styleformer-C, which combines Styleformer and StyleGAN2. We generate low-resolution parts (up to $32 \times 32$) of image using Styleformer Encoder, and the rest of the resolutions parts of image are generated by applying StyleGAN2 block. Fundamentally, in Styleformer-C, the detail of Styleformer part is the same as that of Appendix \ref{app:A}, and the StyleGAN2 part is the same as that of StyleGAN2 \cite{karras2020analyzing} implementation. We experiment with CLEVR and Cityscapes which is high-resolution multi-object or compositional scene datasets with a image size of $256 \times 256$, and we also experiment AFHQ-Cat which is a high-resolution single-object dataset with a image size of $512 \times 512$. We performed all training runs on NVIDIA 2 Tesla V100 GPUs.

\paragraph{CLEVR and Cityscapes}
In CLEVR, we start at $8 \times 8$ learned constant input just like the default setting and used $\{1,2,1,1,1,1\}$ for Layers, $\{1024,256,256,256,256,128\}$ for Hidden size, training for 10M images. Here, what Layers means in StyleGAN2 is a block which has two convolution operations and what Hidden size means in StyleGAN2 is the number of channels. In Cityscapes, we apply StyleGAN2 in the same way as CLEVR. We also set Layers and Hidden size same as CLEVR experiment. Furthermore, we train Styleformer-C for 36M images.

\paragraph{AFHQ-Cat}
Likewise, we start with $8 \times 8$ learned constant input and generate an image of $512 \times 512$ size. We set Layers to $\{1,2,1,1,1,1\}$ and Hidden size to $\{1024,256,256,256,256,64\}$ to train Styleformer-C for 9M images.

\section{Visual Samples}
We show various samples generated with Styleformer, Styleformer-L and Styleformer-C in this section.
\label{app:G}

\begin{figure*}[t]
\begin{center}
\includegraphics[height=1.0\linewidth, width=1.0\linewidth]{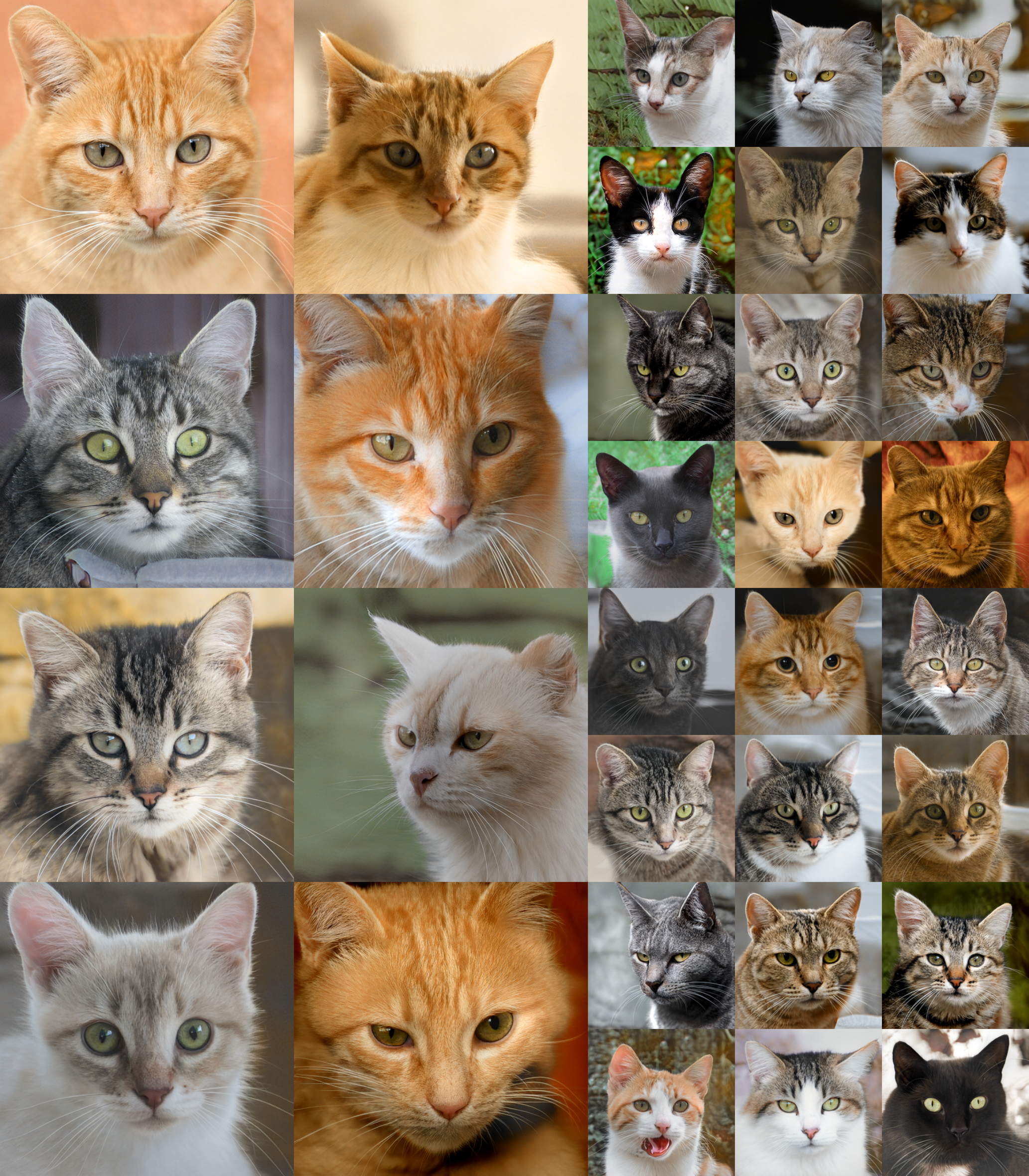}
\end{center}
\vspace{-5mm}
\caption{High-resolution samples generated by Styleformer-C on AFHQ-Cat.
}
\vspace{-3mm}
\label{fig:afhq_app}
\end{figure*}

\begin{figure*}[t]
\begin{center}
\includegraphics[height=1.0\linewidth, width=1.0\linewidth]{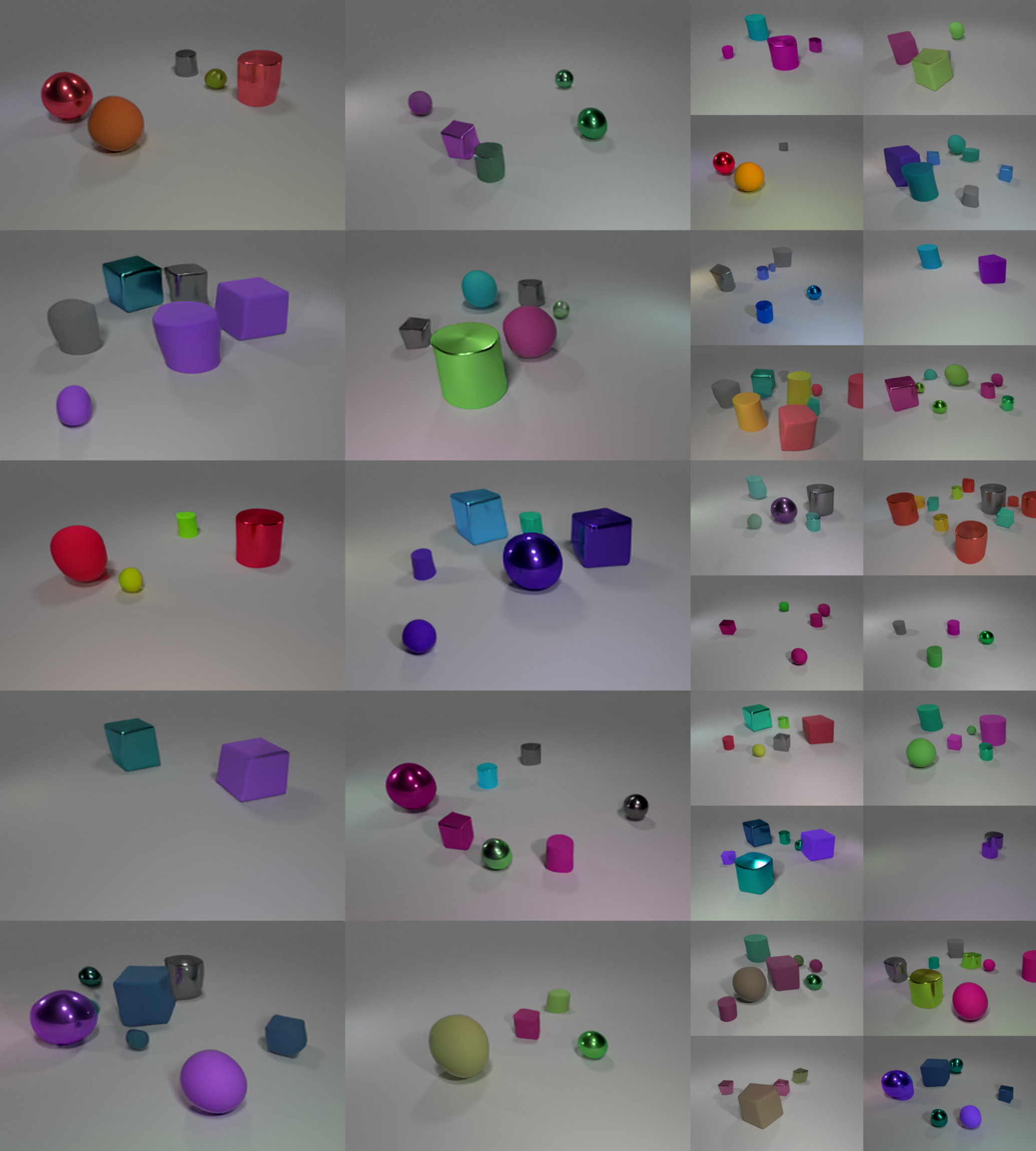}
\end{center}
\vspace{-5mm}
\caption{High-resolution samples generated by Styleformer-C on CLEVR.
}
\vspace{-3mm}
\label{fig:clevr_app}
\end{figure*}

\begin{figure*}[t]
\begin{center}
\includegraphics[height=1.0\linewidth, width=1.0\linewidth]{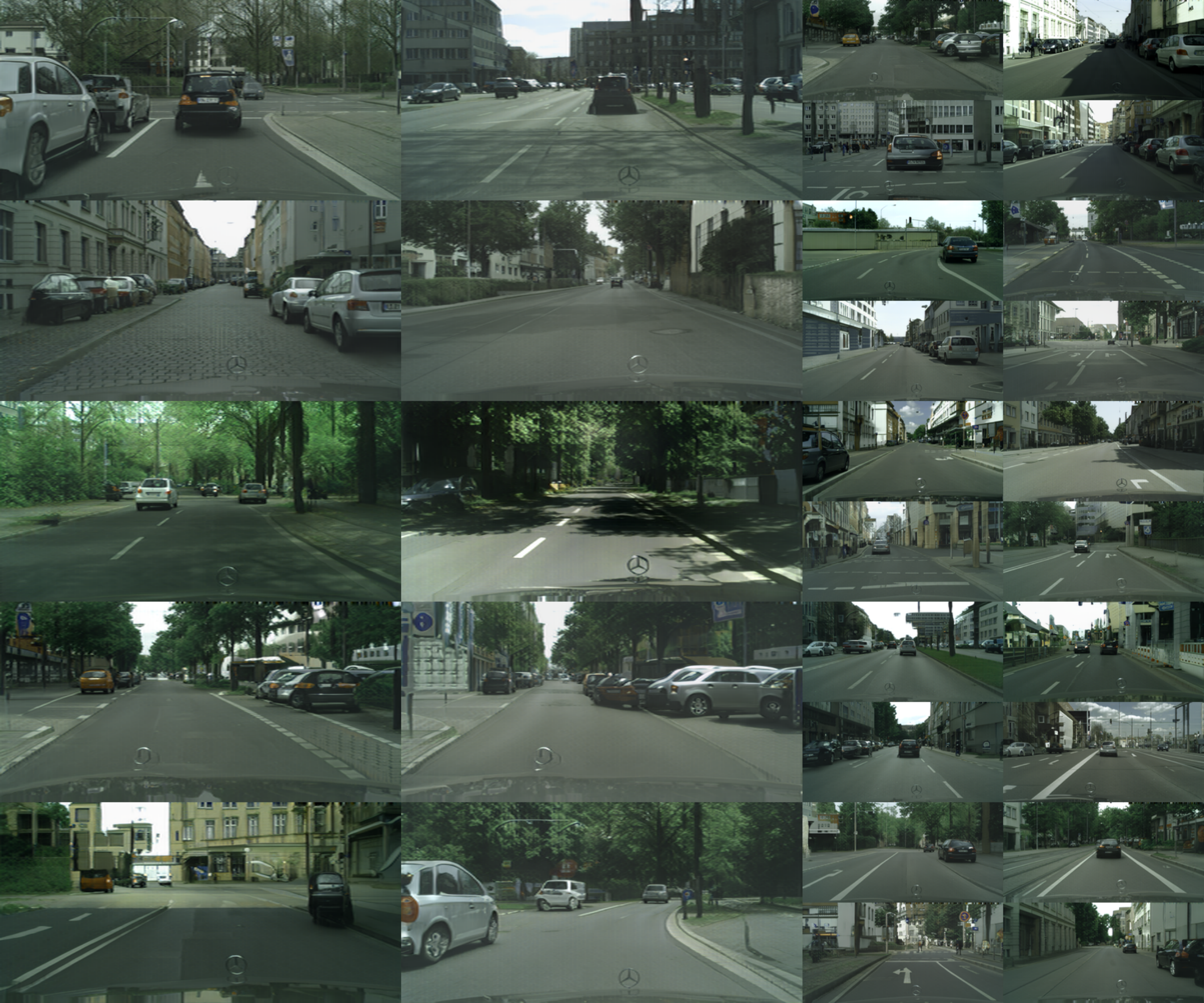}
\end{center}
\vspace{-5mm}
\caption{High-resolution samples generated by Styleformer-C on Cityscapes.
}
\vspace{-3mm}
\label{fig:cityscape_app}
\end{figure*}

\begin{figure*}[t]
\begin{center}
\includegraphics[height=1.0\linewidth, width=1.0\linewidth]{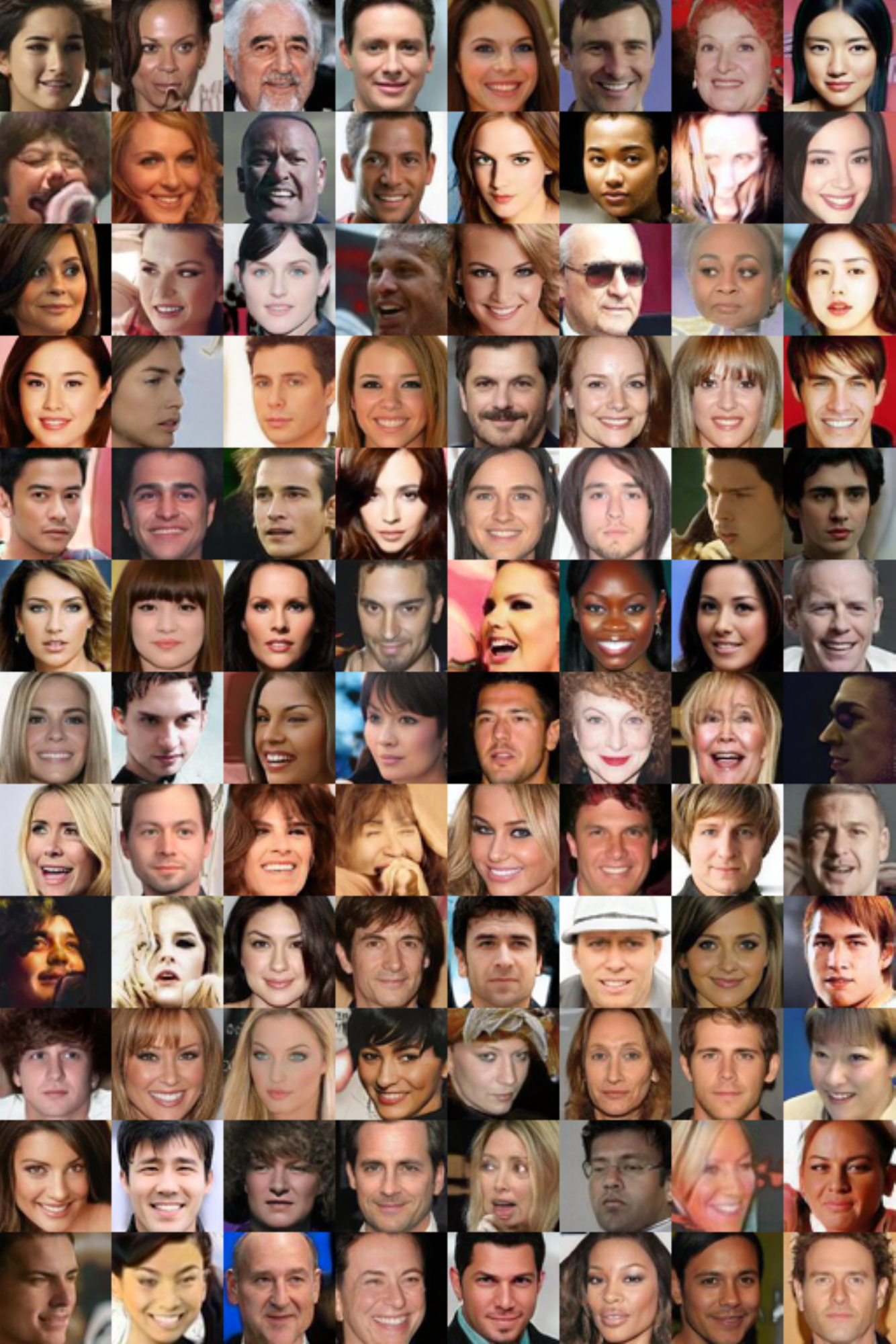}
\end{center}
\vspace{-5mm}
\caption{High-resolution samples generated by Styleformer-L on CelebA.
}
\vspace{-3mm}
\label{fig:celeba_app}
\end{figure*}

\begin{figure*}[t]
\begin{center}
\includegraphics[height=1.0\linewidth, width=1.0\linewidth]{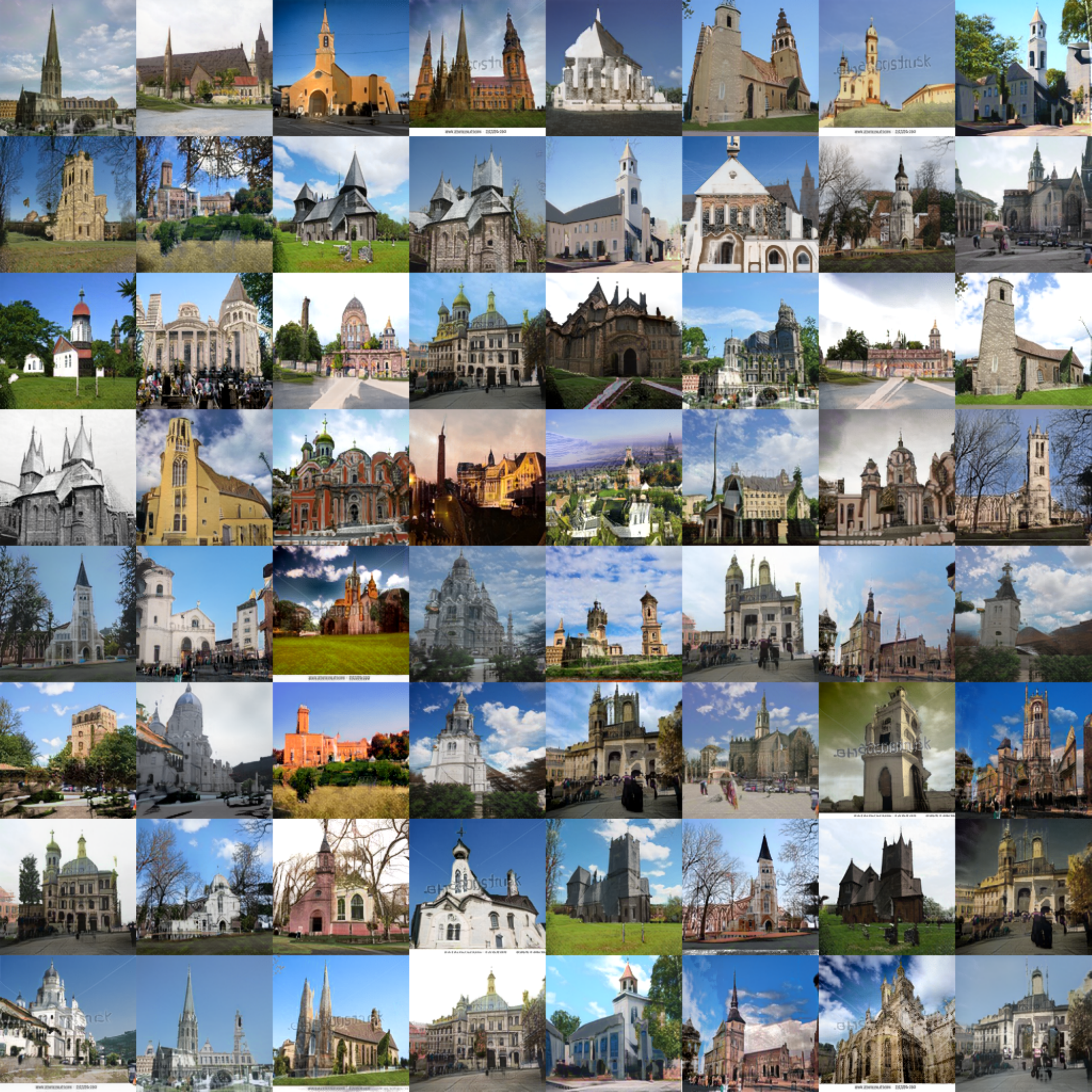}
\end{center}
\vspace{-5mm}
\caption{High-resolution samples generated by Styleformer-L on LSUN-church.
}
\vspace{-3mm}
\label{fig:lsun_app}
\end{figure*}

\begin{figure*}[t]
\begin{center}
\includegraphics[height=1.0\linewidth, width=1.0\linewidth]{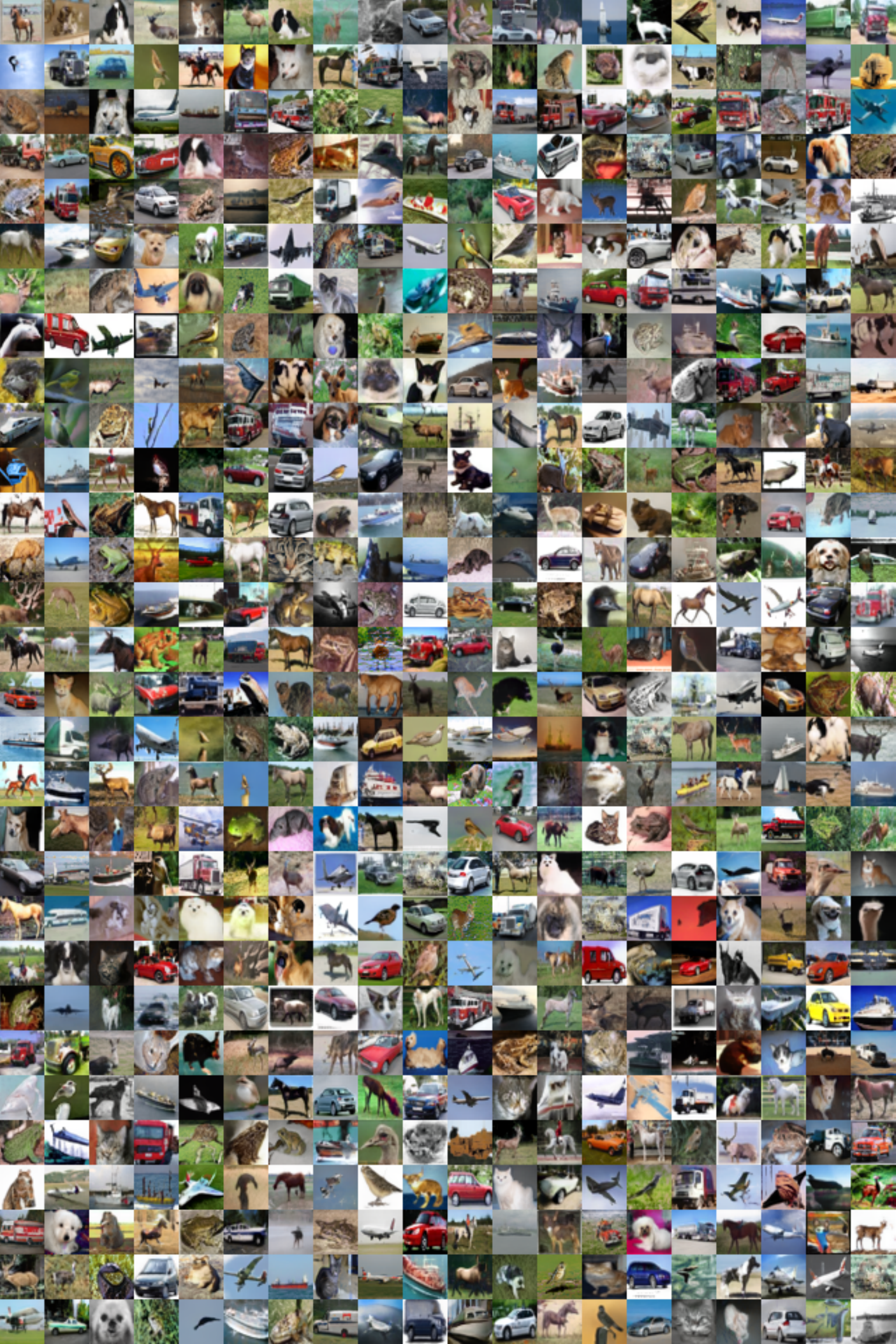}
\end{center}
\vspace{-5mm}
\caption{Samples generated by Styleformer on CIFAR-10.
}
\vspace{-3mm}
\label{fig:cifar_app}
\end{figure*}


\end{document}